\documentclass[twoside,11pt]{article}

\usepackage{jmlr2e}
\usepackage{microtype}
\usepackage{graphicx}
\usepackage{subfigure}
\usepackage{caption}

\usepackage{booktabs} % for professional tables
\usepackage{tikz} % Package for drawing
\usepackage{amsmath}
\usepackage{amssymb} 
\usepackage{flexisym}
\usepackage{graphicx}
\usepackage[utf8]{inputenc}
\usepackage{pgfplotstable}
\usepackage{url} 
\usepackage{multicol}
\newcommand{\rbigan}{MDGAN}
\newcommand{\prbigan}{P-MDGAN}
\newcommand{\prbiganml}{P-MDGAN with ML}
\newcommand{\prbiganr}{EP-MDGAN}

\renewcommand{\vec}[1]{\mathbf{#1}}
\newcommand{\z}{\vec{z}}

\newcommand{\x}{\vec{x}}

\newcommand{\lno}{\lambda_{\text{norm}}}
\newcommand{\Lm}{\mathcal { L } }
\newcommand{\lcyc}{\lambda_{\text{cyc}} }

\newcommand{\ldist}{\lambda_{\text{dist}} }
\newcommand{\lperc}{\lambda_{\text{perc}} }

\newcommand{\logl}{\log_{10} p(G(E(\x)))}

\jmlrheading{}{}{}{}{}{}

\ShortHeadings{Improved BiGAN training with marginal likelihood equalization}{S\'anchez-Mart\'in,  M. Olmos and Perez-Cruz}
\firstpageno{1}
\begin{document}

\title{Improved BiGAN training with marginal\\ likelihood equalization}

\author{\name Pablo S\'anchez-Mart\'in \email pablo.sanchez-martin@tuebingen.mpg.de\\
       \addr Max Planck Institute for Intelligence Systems\\
       T{\"u}bingen, Germany
       \AND
       \name Pablo M. Olmos \email olmos@tsc.uc3m.es \\
       \addr University Carlos III in Madrid\\
       Madrid, Spain
   \AND
   \name Fernando Perez-Cruz \email fernando.perezcruz@sdsc.ethz.ch \\
   \addr Swiss Data Science Center (ETH Z\"urich) \\
   Z\"urich/ Laussane, Switzerland}

\maketitle

\begin{abstract}%   <- trailing '%' for backward compatibility of .sty file

We propose a novel training procedure for improving the performance of generative adversarial networks (GANs), especially to bidirectional GANs. First, we enforce that the empirical distribution of the inverse inference network matches the prior distribution, which favors the generator network reproducibility on the seen samples.
Second, we have found that the marginal log-likelihood of the samples shows a severe overrepresentation of a certain type of samples. To address this issue, we propose to train the bidirectional GAN using a non-uniform sampling for the mini-batch selection, resulting in improved quality and variety in generated samples measured quantitatively and by visual inspection. We illustrate our new procedure with the well-known CIFAR10, Fashion MNIST and CelebA datasets.

\end{abstract}
 
% !TEX root = ./main.tex
\section{Introduction}
\label{sec:intro}

Implicit generative modeling, in general, and Generative Adversarial Networks (GANs) \citep{G1}, in particular, promise to solve the universal simulator problem in an end-to-end fashion \citep{G1, G2, G3}. GANs have been successfully applied to a variety of tasks, such as image-to-image translation \citep{Aimimtrans}, image super-resolution \citep{SRes}, image in-painting \citep{APaint}, domain adaptation \citep{ACrossDom}, text-to-image synthesis \citep{StackGAN}, dark matter estimation \citep{ADarkMatter}, and breaking federated learning systems \citep{Hitaj17}, among many others. Progress in GANs has been quite remarkable and fast in the past four years. Most of the work has concentrated on improving its training to make it more stable, robust and generalizable to numerous architectures and datasets \citep{D1, WGANGP, WGAN,  MMDGAN, SNDCGAN} to name a few. All these approaches focus on learning the unidirectional mapping from a low-dimensional embedding $\z$ to the high-dimensional data space $\x$. Several works in the literature have reported that the latent space of GANs drives semantically meaningful representations \citep{chen2016infogan}. Further, data augmentation and image editing techniques can be built upon the latent space of GANs \citep{MR2, MR3, GANPaint}. 

Another interesting extension to GANs are ALI \citep{ALI} and BiGAN \citep{BiGAN} that propose a novel framework for training bidirectional GANs, that jointly learns the bidirectional mapping between $\z$ and $\x$ in an unsupervised manner. They use the generator similar to unidirectional GANs \citep{MR2, LSGAN, WGANGP, DenGAN} for constructing the forward mapping from $\z$ to $\x$, and then use an encoder network to model the inference mapping from $\x$ to $\z$. The discriminator network is now trained to distinguish the joint distribution of $(G(\z), \z)$ from that of $(\x, E(\x))$, where  $E(\cdot)$ and $G(\cdot)$ respectively denote the mapping functions defined by the encoder and the generator. Although their models can reconstruct the original image from the estimated latent variable, the visual quality of the generation is generally worse than the unidirectional GANs.  MDGAN \citep{MDGAN}, unlike ALI and BiGAN, introduces an additional constraint to enforce that the reconstructed images are identical to the original images, namely the reconstruction loss. VEEGAN \citep{VEEGAN} also includes a reconstruction loss on the latent space to also improve reconstruction quality.

In this paper, we address several technical challenges not yet reported for (bidirectional) GANs. First, while the reconstructed images $G(E(\x))$ by MDGAN are accurate, we notice that the empirical distribution induced by $E(\x)$ at the latent space $\z$ for both train and test images is not from the typical set of the latent space distribution\footnote{This issue also appears in standard BiGAN.}. For example, if $p(\z)$ is a $D$-dimensional Gaussian, then the induced $||E(\x)||_2$ from MDGAN training is typically much larger than $\sqrt{D}$, thus implying that the real images are not typical from the latent distribution standpoint. To address this issue, we include in the loss function a second regularization term that penalizes $||E(\x)||_2\neq\sqrt{D}$. 

Second, once the encoder maps the training (and test) images to the typical set of the latent space, we study the marginal log-likelihood per image in both the train and test sets. We noticed that the log-likelihood distribution  is extremely asymmetric, this issue was not raised by \citep{LR2}, in which the marginal likelihood was first computed for GANs/VAEs. To compensate for this effect, we propose to train MDGAN using two non-uniform sampling schemes. In the first scheme, we use the estimated marginal log-likelihood to reweigh the training samples probability of being added to each mini-batch. In the second approach, the reweighing uses the reconstruction quality as a proxy to measure the marginal log-likelihood. The use of reconstruction quality instead of marginal likelihood was proposed upon observing that both quantities are highly correlated, i.e., overrepresented images are typically reconstructed with higher quality, and computing log-likelihood is computationally expensive, while reconstruction quality can be easily measured after each mini-batch has been trained. We found that this second approach improves both the quality of the generated images, and the diversity of the generative model.
% !TEX root = ./main.tex

\section{Related Work}
\label{sec:related_work}

\subsection{BiGAN and MDGAN}

Currently, there exist several variants GANs that allow performing inference over the latent variable. One of these models is the Bidirectional Generative Adversarial Network (BiGAN) \citep{BiGAN} which includes an inference or encoder network denoted as $E(\x)$ that learns a mapping from data points $\x$ into the latent space $\z$. Now, the discriminator aims at distinguishing between real tuples $(\x, E(\x))$ and fake tuples $(G(\z), \z)$, thus it is denoted as $D(\x, \z)$. The optimization of the model reads as follows:

\begin{equation}
\begin{split}
\label{eq:bigan_obj}
\Lm_{ \mathrm {BiGAN} } &:=\mathbb{E}_{\mathbf{x} \sim p_{\mathbf{X}}}\left[\mathbb{E}_{\mathbf{z} \sim p_{E}(\cdot | \mathbf{x})}[\log D(\mathbf{x}, \mathbf{z})]\right] +\mathbb{E}_{\mathbf{z} \sim p_{\mathbf{Z}}}\left[\mathbb{E}_{\mathbf{x} \sim p_{G}(\cdot | \mathbf{z})}[\log (1-D(\mathbf{x}, \mathbf{z}))]\right]
\end{split}
\end{equation}

\cite{BiGAN} demonstrate that at the global optimum, \textit{G}  and \textit{E} are each other's inverse. Unfortunately, in practice we are only able to arrive at a local optimum. As a consequence, the reconstruction of a sample,  $G(E(\x))$, has margin for improvement. \cite{MDGAN} propose to add a reconstruction loss term
\begin{align}
\label{eq:rbigan_rec}
\Lm_ { \mathrm { cyc } } ( G , E ) = \mathbb { E } _ { \x \sim p _ { r } ( \x ) } [ d(G(E(\x)), \x)] 
\end{align}

to the original objective function in order to drive  better reconstructions. This term measures the reconstruction error based on some distance metric $d(\cdot,\cdot)$, typically minimum squared error. They named the resulting model MDGAN. The updated  objective function is $\Lm ( D , G, E ) =  \Lm_{ \mathrm {BiGAN} } +   \lcyc \Lm_ { \mathrm { cyc } } ( G , E ) $, where $ \lcyc$ is a new hyperparameter controlling the reconstruction quality. Results show a clear improvement in the reconstructed data points, as expected. 

\subsection{Data log-likelihood in GANs}

 The likelihood of a test sample can be computed as follows:
 \begin{equation}\label{like}
 	p(\x_{\text{test}})=\int p(\x_{\text{test}}|\z)p(\z)d\z,
 \end{equation} 
 In \cite{LR2}, the authors proposed an isotropic Gaussian likelihood for GANs, i.e:
 \begin{equation}\label{aprox_LR2}
 	p(\x_{\text{test}}|\z)\approx\frac{1}{(2\pi\sigma^2)^{\dim(\x)}}\exp\left(\frac{||\x_{\text{test}}-G(\z)||^2}{2\sigma^2}\right).
 \end{equation}
 They solved the integral in \eqref{like} by annealed importance sampling \citep{neal2001annealed}. This model for $p(\x_{\text{test}}|\z)$ is a fine choice, if the error in the generated images comes from noisy observations and there is not a significant model misspecification. But when the error in GAN generation comes from model misspecification (i.e. there is little noise in images datasets) finding the appropriate $\sigma$ becomes an impossible task, because not all samples will be equally represented by the model. In this case, there is no right $\sigma$ that helps us evaluate the marginal likelihood accurately for our model. When the reconstruction can be uneven, best reconstructed images would seem more likely, which does not need to be the case. Hence, the estimated likelihood can be biased by the sample's reconstruction quality in a way that does not allow distinguishing which one is at play. 

More recently, \cite{EnGAN}  proposed an approach to compute data log-likelihood for optimal transport GANs, e.g. Wasserstein GAN \citep{WGAN}, based on three main components: the distance between real samples and the generative model, the entropy of the posterior distribution and the likelihood of the coupled latent variable. They also propose a modification of their approach to be used in a general GAN by not taking into account the entropy term. 

 \begin{figure}[t!]
	\centering
	\subfigure[\rbigan]{\includegraphics[width=0.5\linewidth]{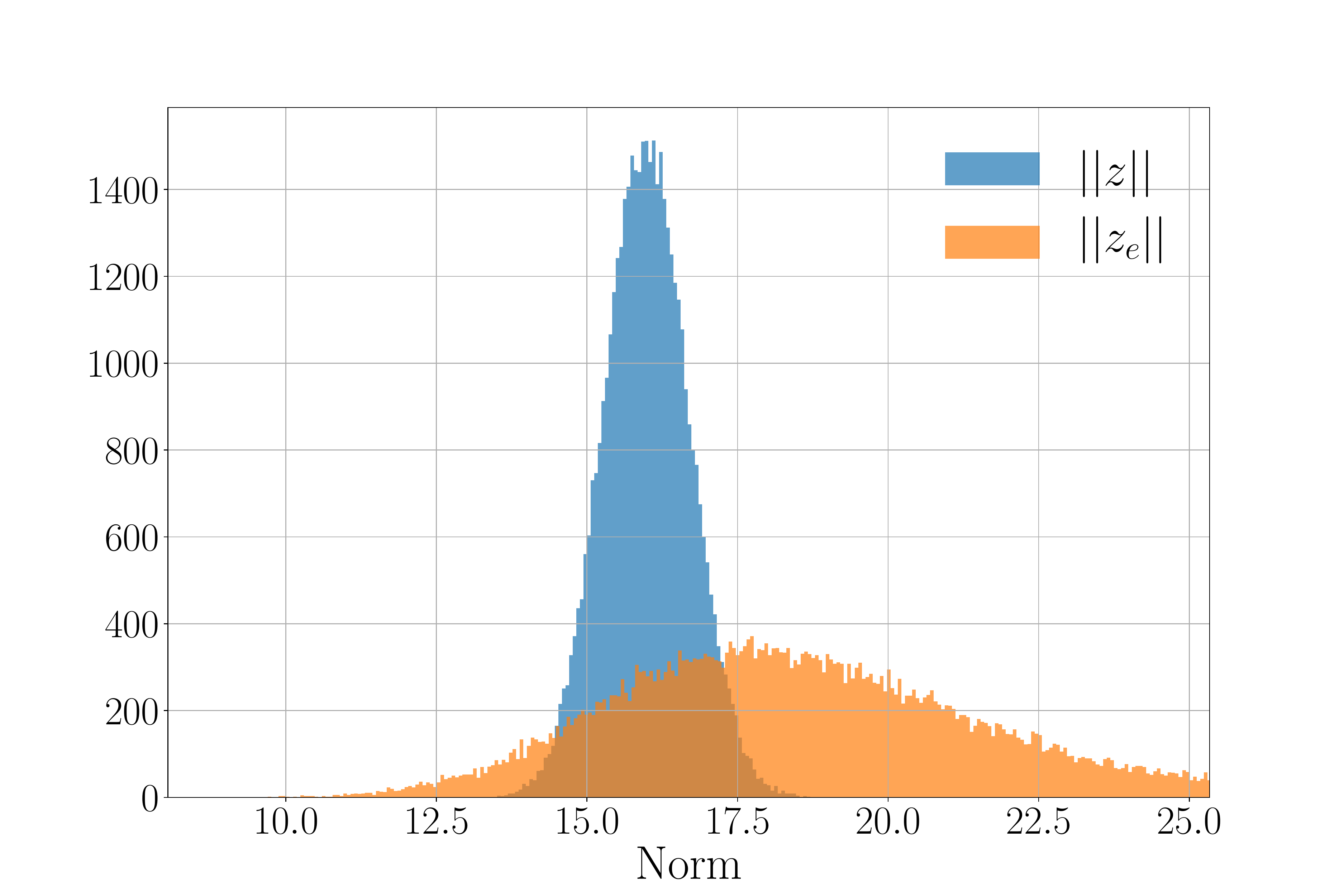}}%
	\subfigure[\prbigan]{\includegraphics[width=0.5\linewidth]{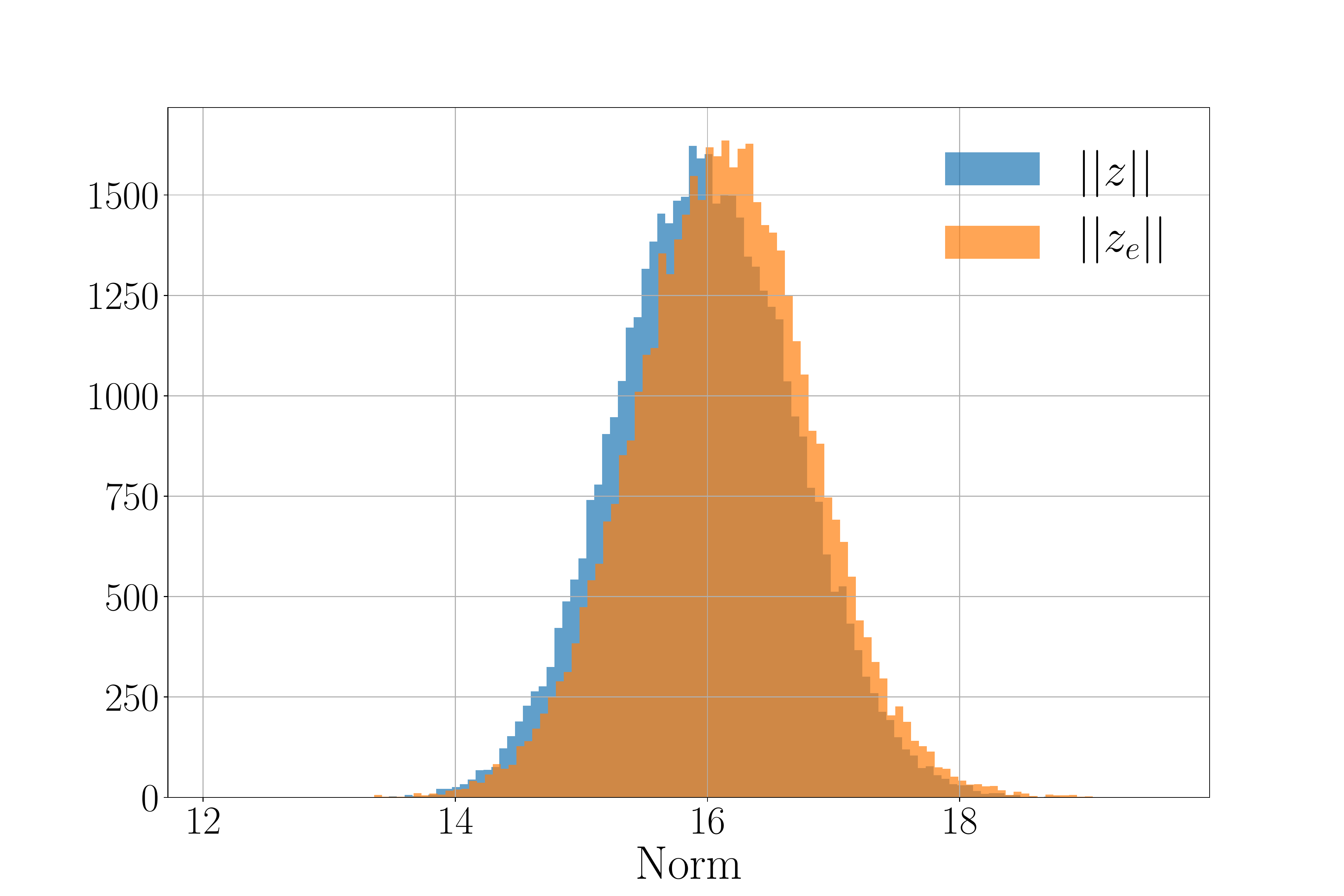}}%	
	
	\caption{Empirical distribution of $||\z||_2$ for prior samples and  for  encoded vectors $\z_e = E(\x)$ computed over the CIFAR10 train set for \rbigan\ (a)  and \prbigan\ (b).}%
	\label{fig:z_emp_dist}
	\vskip -0.05in
\end{figure}

% !TEX root = ./main.tex

\section{Prior Regularized \rbigan}
\label{sec:PR-BiGAN}

While \rbigan\ is designed to provide realistic reconstruction of the original image $\x \approx G(E(\x))$, it ignores the fact that latent projections produced by $E(\x)$ can be untypical under the latent-space model $p(\z)$, where $p(\z)$ is usually taken as an independent high-dimensional Gaussian distribution (or other simple distribution).

For example, we have trained an \rbigan\ with a 5-layer generator network, a 7-layer discriminative network and a 256-dimension latent space (see the specific network structural details in Appendix \ref{app:architecture}) with the images from CIFAR10. We have taken the 45,000 training images and we have recovered the latent space representation $\z_i=E(\x_i)$.  In Figure \ref{fig:z_emp_dist} (a), we plot the histogram of the norm of these latent space representation (in orange). For comparison, we plot the histogram of the norm of 45,000 samples from $p(\z)$ in blue.  Observe that there is an important fraction of images for which $||\z_i||$ is outside the typical set of $p(\z)$ and they will not be generated when sampling from $p(\z)$, since the prior probability mass decays exponentially fast with $||\z||^2$. To address this issue, we propose to add a simple additional regularizer to the \rbigan\ objective function:
\begin{align}
\label{eq:approach4_reg}
\Lm_ {\mathrm { norm }} ( E ) = \mathbb { E } _ { \x \sim p _ { r } ( \x ) } \left[  \left(|| E(\x)||_2-\sqrt{\text{dim}(\z)} \right)^2 \right],
\end{align}
which penalizes values of the norm of  $E(\x)$ that lay outside the typical set. For other prior distributions, this regularizer should be adjusted accordingly. Finally, the objective function of this new scheme, we refer to as Prior-regularized \rbigan\ (P-MDGAN), is:
\begin{align}
\label{eq:PR-BiGAN}
\Lm ( D , G, E ) =  \Lm_{ \mathrm {BiGAN} }  + \lambda_{\text{cyc}} \Lm_ { \mathrm { cyc } } +  \lambda_{\text{norm}}  \Lm_ { \mathrm { norm } },
\end{align}
where $ \lambda_{\text{norm}} $ is an automatically-tuned hyperparameter during training that ensures the distribution of the encoded latent vectors matches the prior distribution (details can be found in Appendix \ref{app:training_details}). In the Figure \ref{fig:z_emp_dist} (b), we show the norm of the latent space representation of the same 45,000 training images used for MDGAN, in which we can see that the \prbigan\ latent space ensures the samples are mapped to the typical set.

% !TEX root = ./main.tex

\section{Equalized \rbigan\ training}
\label{sec:ml_scheme}

In this section, we describe our second contribution of this paper. We first note that the marginal likelihood and reconstruction quality between the original images and the reconstructed images varies orders of magnitude and we propose to adapt the training of GANs to try to equalize both. In Section \ref{sec:related_work}, we presented two methods to compute the marginal likelihood that mixed in a single number both the marginal likelihood and reconstruction quality. In this paper, we also advocate for measuring both of them independently, because we believe that each one of them provide us with different information that would be useful to improve the performance of bi-directional GANs. In this paper, we also go one step further and propose to use these measurements to modify the training and as consequence improve the performance of the resulting GAN.

\begin{figure}[t!]
	\centering
	\begin{minipage}{0.48\textwidth}
		\centering
		\vspace{-1cm}
		\includegraphics[width=\textwidth]{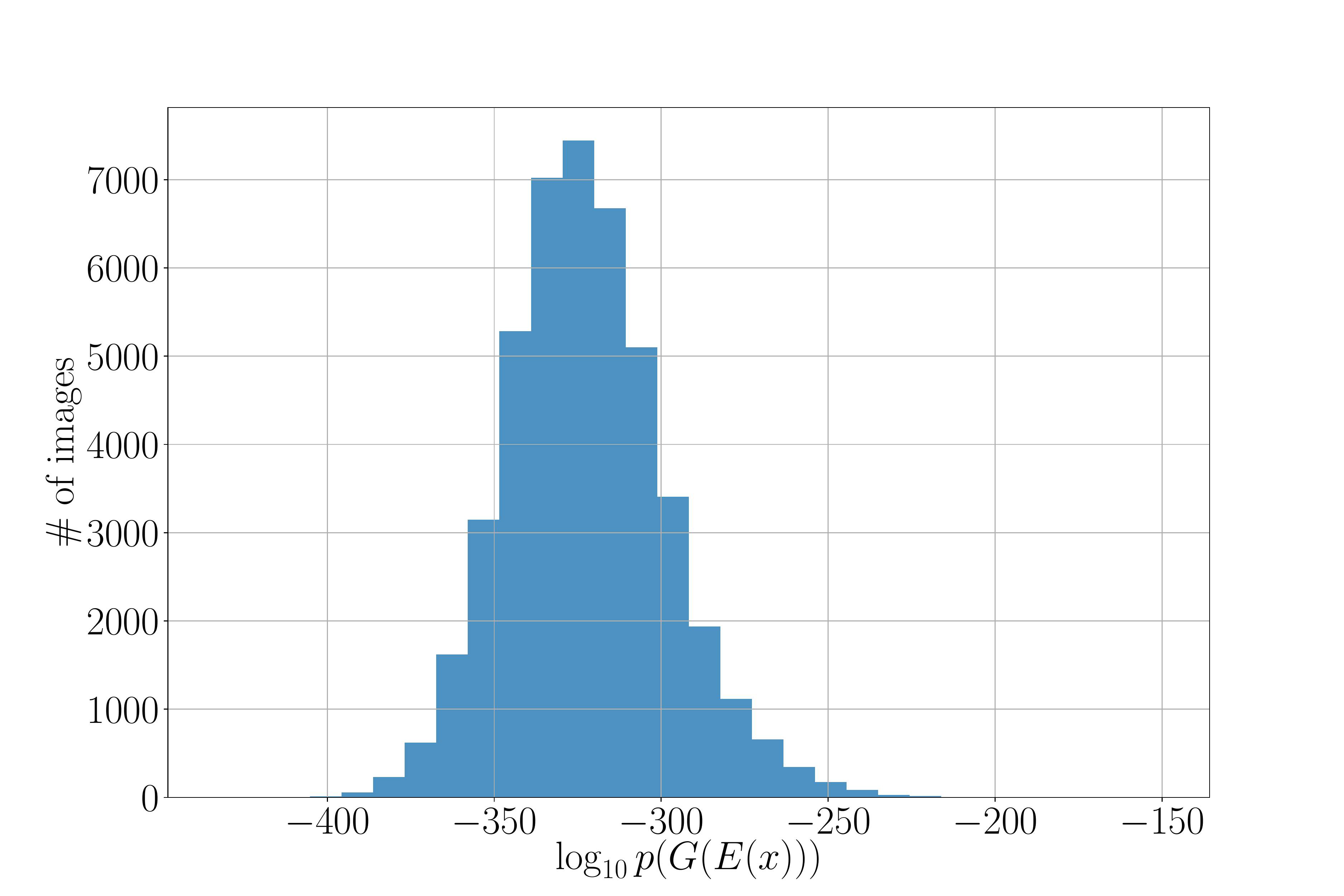}%
		\vspace{0.5cm}
		\caption{Empirical distribution of $\logl$ for all CIFAR10 training examples.}%
		\label{fig:ll_hist}
	\end{minipage}\hfill 
	\begin{minipage}{0.48\textwidth}
	\centering
		\includegraphics[width=\linewidth]{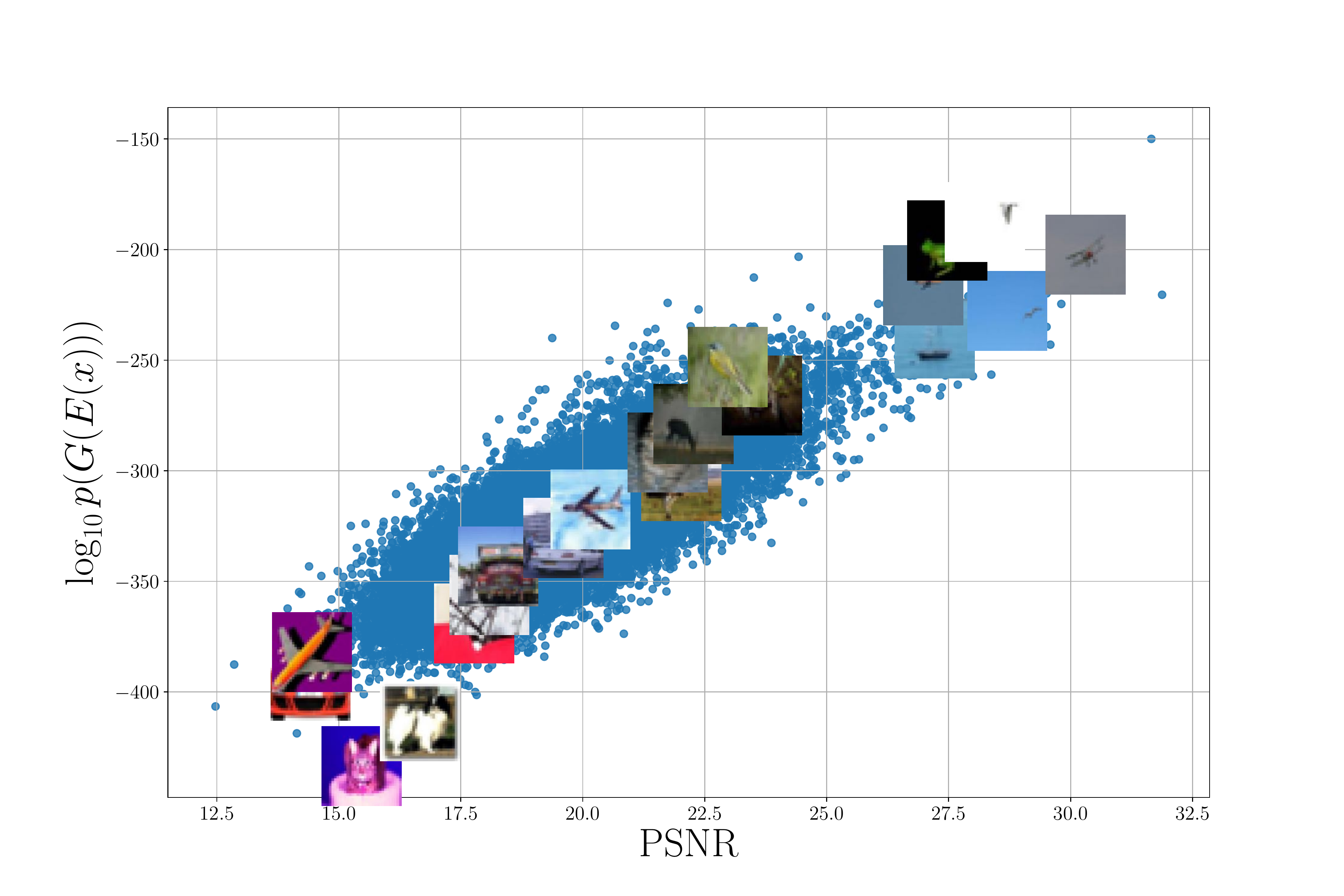}%
		\vspace{0.5cm}
	\caption{Scatter plot of reconstruction quality measured in PSNR versus $\logl$ for all examples in the CIFAR10 training set with several original images overlapped.}%
	\label{fig:ll_psnr_scatter}
\end{minipage} 

\end{figure}

Instead of estimating the marginal likelihood in \eqref{like}, we propose to compute the marginal likelihood of the reconstructed image, i.e.:

 \begin{equation}\label{like2}
p(G(E(\x)))=\int p(G(E(\x))|\z)p(\z)d\z,
\end{equation}
because in this way we know how likely it is the generated image from a given $\z$, independent of the reconstruction quality. We also measure the quality of the reconstructed image using the mean squared error, as proposed in the original MDGAN paper to enforce reconstruction quality \citep{MDGAN}. We actually report the Peak Signal-to-Noise Ratio (PSNR):
\begin{align*}
\text{PSNR}(\x,G(E(\x))) = 10 \log_{10}\left( \frac{3\times K\times M^2}{||\x - G(E(\x))||^2} \right),
\end{align*}
where 3 is the number of color channels, $K$ is the number of pixels in the image and $M$ is the maximum possible pixel value of the images, i.e. 255 for 8-bit color images.

For illustration purposes, we consider a \prbigan\ trained over CIFAR10, as in the previous section. In Figure \ref{fig:ll_hist}, we report the log-likelihood of the 45,000 reconstructed training images computed using a similar procedure to \cite{LR2} for \eqref{like2}. Considering that every image was used equally during training, we would expect that they all have similar log-likelihoods, yet we observe the differences are quite significant. The values shown are in $\log_{10}$, this means the differences are in the order of $10^{100}$ between the most generated images and the least generated images.

These results indicate that the \prbigan\ is substantially over-representing a small subset of images, which are orders of magnitude more likely than the rest of the images. Figure \ref{fig:ll_psnr_scatter} shows a  scatter plot comparing the reconstruction quality of $G(E(\x))$ w.r.t. the original image $\x$ measured in PSNR versus their corresponding estimated marginal log-likelihood, we can observe that images with simpler textures and large uniform backgrounds are not only reconstructed with better quality but also they are being overrepresented by the generator network. In parallel to our work, \citep{krusinga2019understanding} also acknowledged this phenomenon.

In order to improve the diversity and reconstruction quality of the generated/reconstructed images, we propose to retrain the \prbigan\ from scratch using non-uniform sampling. We rely on the marginal log-likelihood estimation from the first \prbigan\ to boost the marginal likelihood and reconstruction quality of the less represented samples. The main idea is to resample more frequently in the mini-batches those samples with lowest marginal likelihood. We use a simple scheme that shows significant improvements\footnote{This is a first attempt to show that non-uniform sampling is useful to train GANs, further research is needed to find an optimal resampling strategy.}, In particular, we first sort the images according to their $LL(G(E(\x_k)))=\log_{10} p(G(E(\x_k)))$ from larger to smaller, their probability in the non-uniform distribution will be proportional to the position $k\in[1,2,\ldots,N]$ they hold in the sorted list, raised to the power of $\ldist\geq0$. Additionally, we introduce a second hyperparameter, $\lperc\in[0,1]$, to control the percentage of samples per mini-batch that use this re-weighting strategy, while the remaining are chosen uniformly. We refer to this non-uniform sampling scheme as P-MDGAN with MLeq, where MLeq stands for Marginal Likelihood equalization training. The probability of a sample being added to each mini-batch is given by: 
\begin{align}
\label{eq:non_uniform}
 Pr(\x_k | LL(G(E(\x_k))), \ldist, \lperc)= \frac{1-\lperc}{N} +\lperc\frac{\left(k/100 \right)^{\ldist}}{\sum_{j=1}^{N} \left(j/100 \right)^{\ldist}}.\notag
\end{align}

\begin{figure}[h!]
	\centering
	\includegraphics[width=0.65\linewidth]{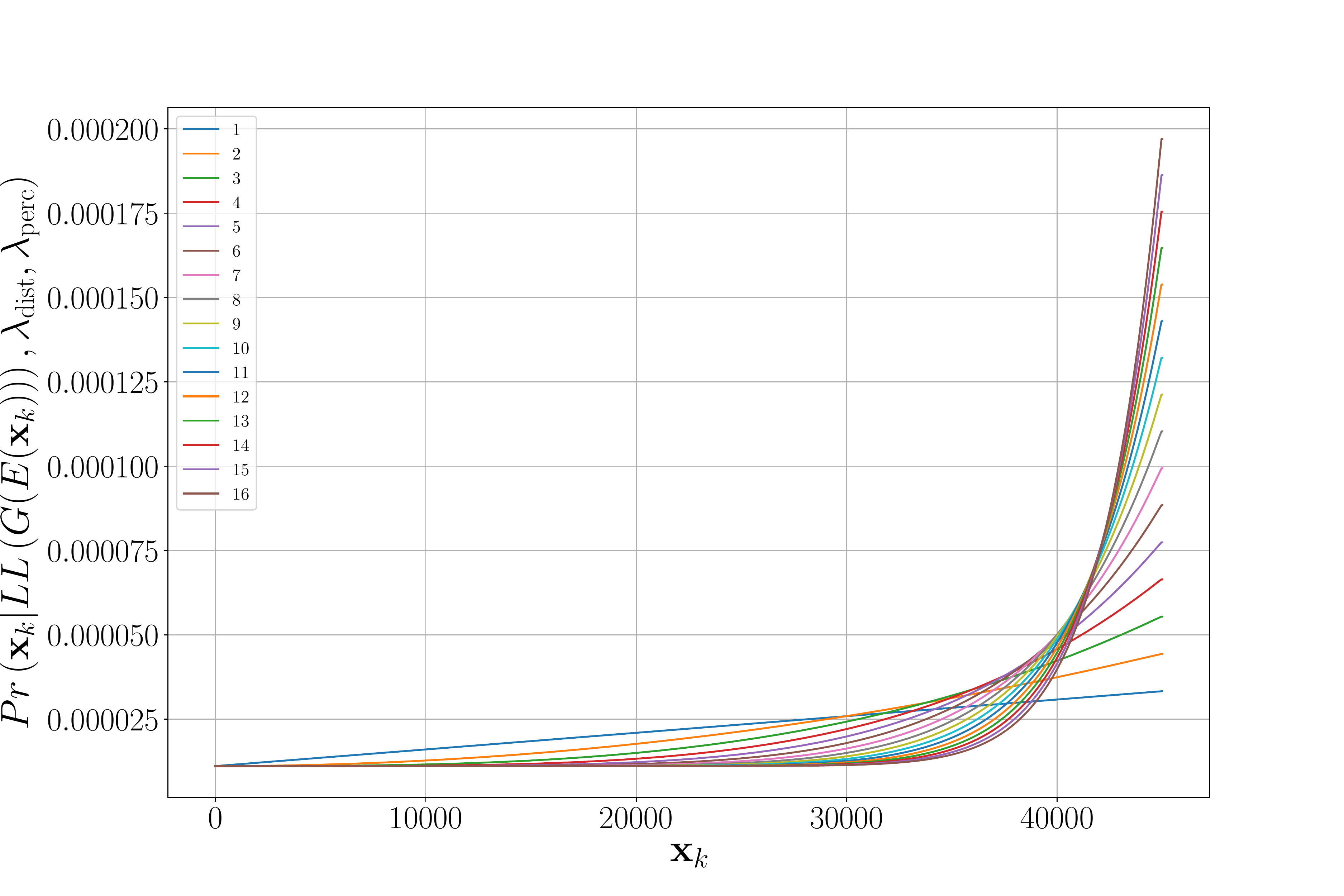}%
	\caption{Non-uniform distribution for different values of $\ldist$.}%
	\label{fig:probs}
\end{figure}

Figure \ref{fig:probs} shows the resulting distribution for different values of $\ldist$.

Our experimental results show that this resampling strategy improves the sample diversity of the generative model (see Section \ref{sec:experiments}). Nonetheless, this training procedure is computational expensive considering we must compute the marginal log-likelihood for all the training images and perform a second round of training. Since we have observed that there exists an almost linear relation between the marginal log-likelihood and the PSNR (See Figure \ref{fig:ll_psnr_scatter}), we propose to use the PSNR as a proxy for the marginal log-likelihood to decide the reweighing of the samples into each one of the mini-batches. Computing the PSNR is quite inexpensive, and we can hence recompute the PSNR for each image after each mini-batch has been trained and use the new PSNR to decide which samples should be added to the next mini-batch. We use the same weighting function as in \eqref{eq:non_uniform}, replacing $LL(G(E(\x_k)))$ by $\text{PSNR}(\x_k,G(E(\x_k)))$ and sorting the samples according to their PSNR. In this way, we do not need to do the training procedure twice, but just have a single procedure that ensures diversity and quality are improved in every step. We refer to this procedure as Equalized Prior-regularized  MDGAN, or \prbiganr, which incorporates the prior regularization from the previous section to make sure all the samples are matched to the typical set and the PSNR equalization to improve sample diversity and reconstruction quality.
% !TEX root = ./main.tex

\section{Experiments}
\label{sec:experiments}

\subsection{Experimental Setup}

We have conducted a broad set of experiments in order to empirically demonstrate the validity of  the regularization techniques proposed.  Here, we present the results obtained with the models trained over the CIFAR10, F-MNIST and CelebA datasets with the configuration of hyperparameters that yield to the best FID scores, shown in Table \ref{table:config_models}. We split the CIFAR10 dataset in 45,000 training, 5,000 validation, and 1,0000 test examples; the F-MNIST contains 54,000 training, 6,000 validation, and 1,0000 test examples; and the CelebA contains 180,540 training, 20,060 validation, and 1,999 test examples.
% lsun \citep{yu15lsun}  and the LSUN consists of 81891 training, 4550 validation, and 4550 test examples. 
Appendix \ref{app:training_details} contains further details about the hyperparameter cross-validated and Appendix \ref{app:results} contains further results.

\begin{table}[t!]
	\centering
	\begin{tabular}{c|c|c|c} 
		
		\hline\hline
		&$\lcyc$& $\lperc$& $\ldist$ \\
		\hline \hline
		\rbigan& 8 (9) & - & - \\
		\prbigan& 7 (9)& - & - \\
		\prbiganml& 5 (5)  & 0.5 (0.8) & 8 (8)  \\
		\prbiganr& 3 (3) & 0.5 (0.8)& 12 (4)\\
		\hline\hline
	\end{tabular}
	\caption{Configuration of hyperparameters for CIFAR10 and F-MNIST. F-MNIST are denoted in parenthesis.}
	\label{table:config_models}
\end{table}

\paragraph{Metrics: } We measure the FID score, which is considered the state-of-the-art measure for GANs. It provides us a way to identify poor performance but being a single scalar metric is unable to distinguish between sample quality and distribution coverage. We also rely on \cite{PRD_improved} to distinguish between both type of failures. \cite{PRD_improved} proposes a 2-dimensional metric based on the notion of precision and recall that matches our understanding of sample quality and distribution coverage. We also inspect the quality of the reconstructed images measured in PSNR, described in Section \ref{sec:ml_scheme}. Finally, we visually  show how our method improves variety in the generated samples.

\paragraph{Experimental setup: } We use the \rbigan\ as the baseline and compared it with the models we proposed: \prbigan\ and \prbigan\ with the two non-uniform sampling schemes, namely \prbigan\ with MLeq and EP-MDGAN. We only computed \prbigan\ with MLeq for the CIFAR10 dataset because it is computationally expensive and \prbiganr\ achieves similar performance. We trained all the models for 800 epochs using the Adam optimizer \citep{kingma2014adam}, stochastic gradient descent with a batch size of 128 and we use spectral normalization GANs \citep{SNDCGAN}. Additionally, we cross-validated the regularization parameters $\lcyc$, $\ldist$, $\lperc$  to have a complete understanding of their effect on the outcome. For all the models, we defined their three elements, namely $G(\cdot)$, $D(\cdot)$ and $E(\cdot)$, as convolutional neural networks with 5, 10 and 7 layers respectively. We consider a Gaussian noise model for the latent variable  $\z \sim \mathcal{N}(\mathbf{0},\mathbf{I})$ with $\text{dim}(\z)= 256$. 

To facilitate reproducibility of our results, in Appendix \ref{app:architecture} and Appendix \ref{app:training_details} we provide an exhaustive description of the networks architecture and parameters selected regarding the training process. PyTorch implementation for the models is available at \url{https://github.com/psanch21/imp_bigan}.

\begin{figure*}[t!]
	\begin{center}
		\begin{tabular}{cc}
			\subfigure[CIFAR10]{\includegraphics[width=0.45\linewidth]{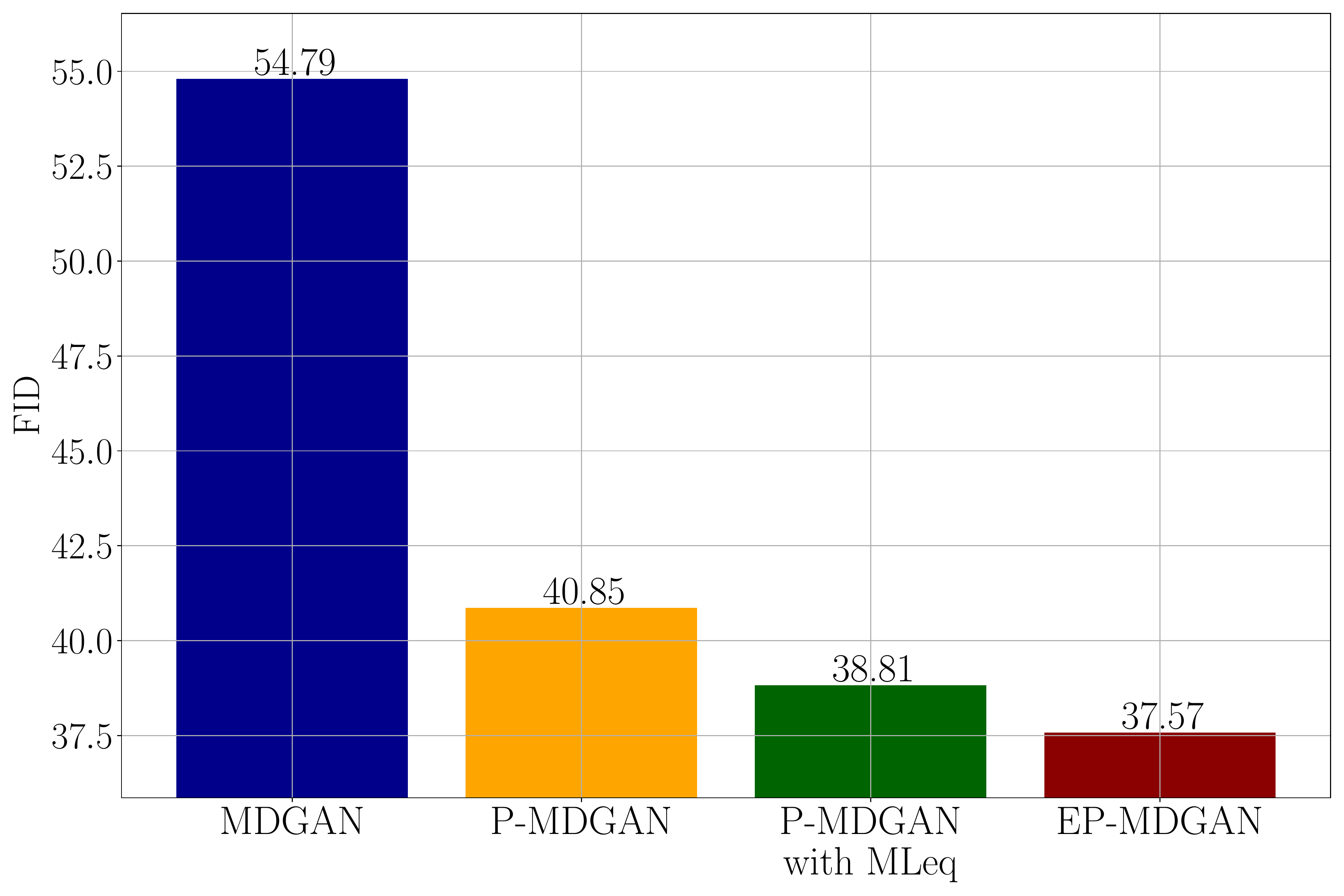}}%
			\hspace{5mm}
			\subfigure[F-MNIST]{\includegraphics[width=0.45\linewidth]{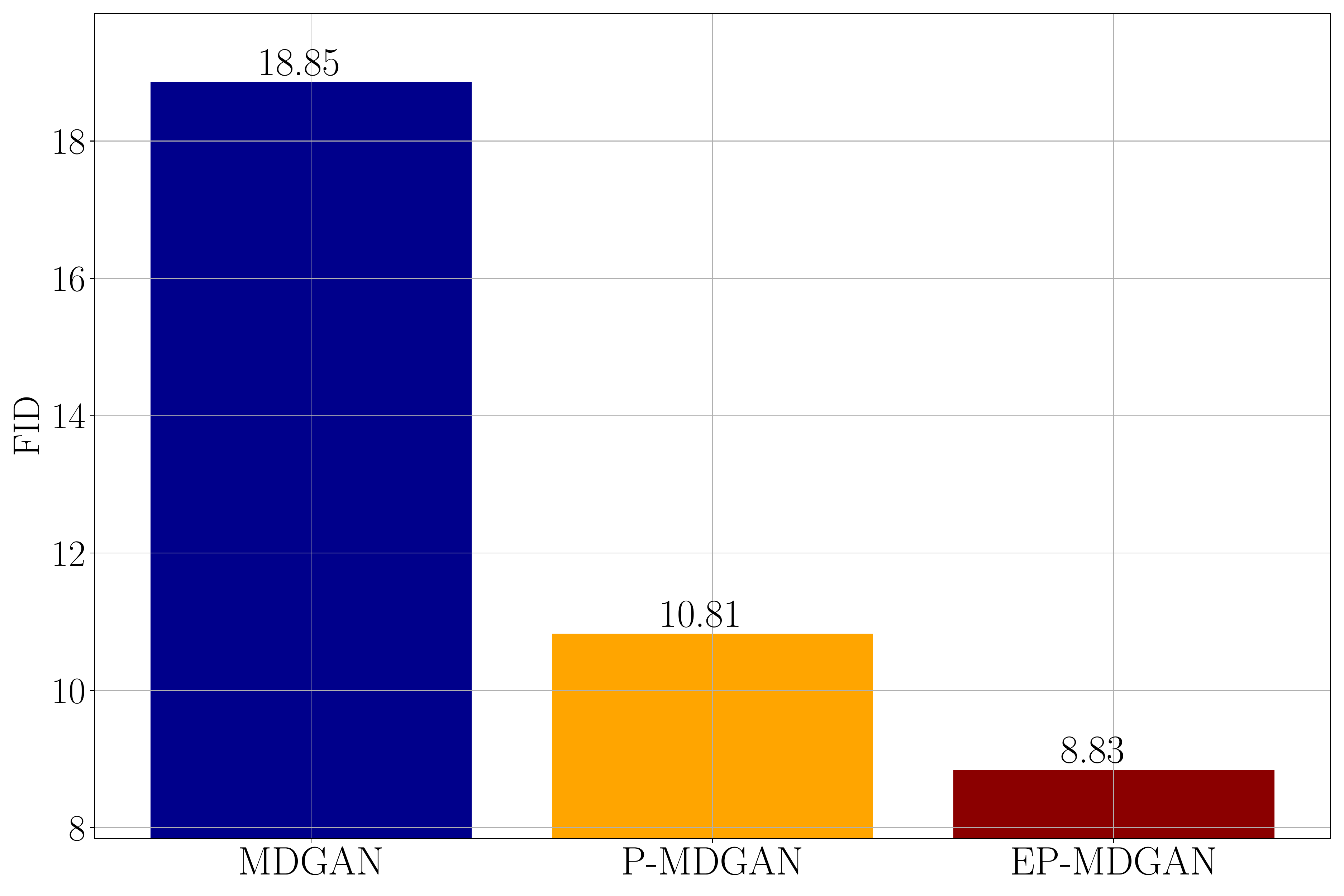}}%
		\end{tabular}
	\end{center}
	\caption{FID scores obtained for 5,000 CIFAR10 and F-MNIST samples from the test set. From left to right: \rbigan, \prbigan,  \prbiganml\  and \prbiganr.}%
	\label{fig:fid_bar}
	\vskip -0.05in
\end{figure*}

\begin{figure*}[t!]
	\begin{center}
		\begin{tabular}{cc}
			\subfigure[CIFAR10]{\includegraphics[width=0.45\linewidth]{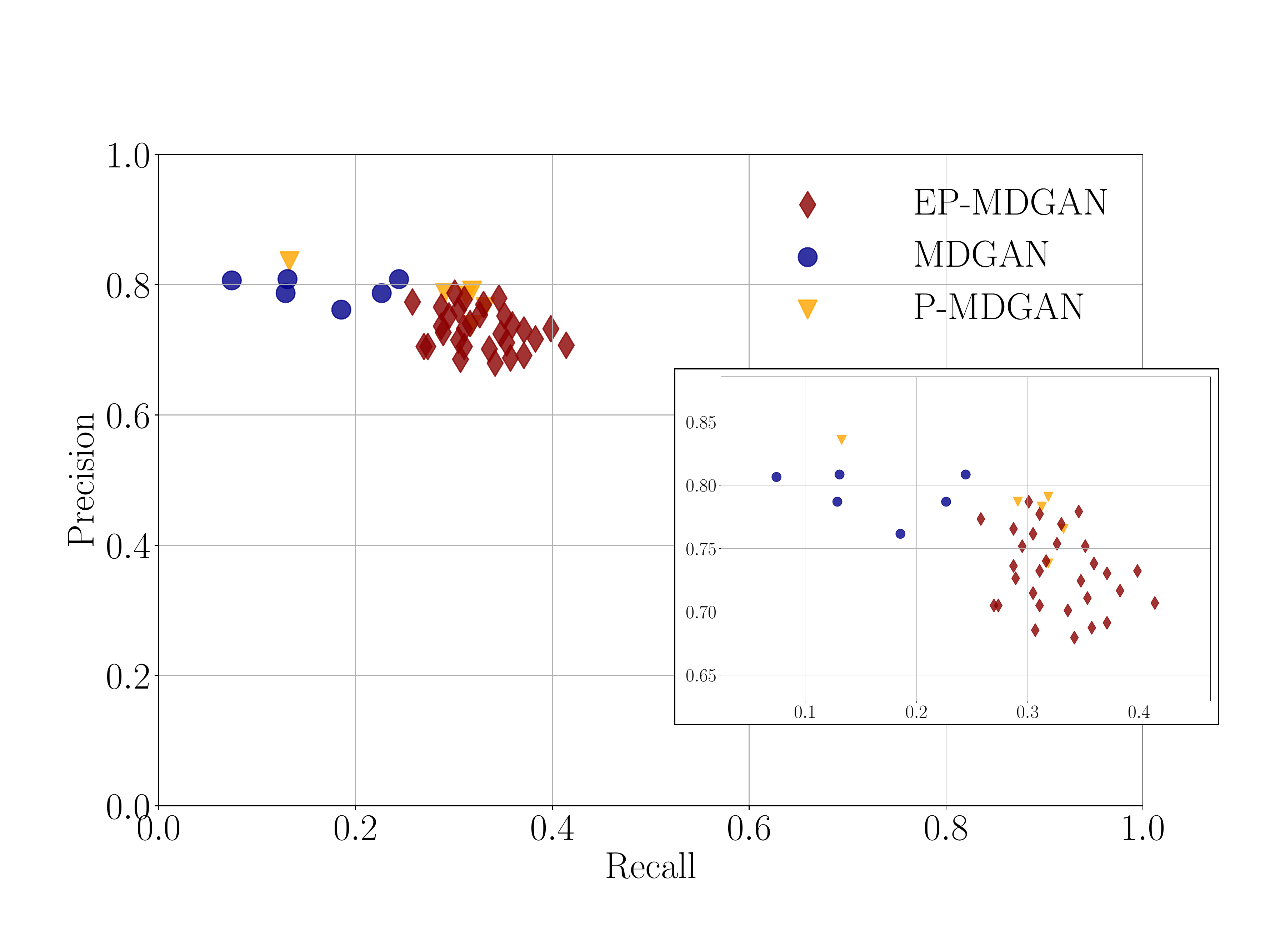}}%
			\hspace{5mm}
			\subfigure[F-MNIST]{\includegraphics[width=0.45\linewidth]{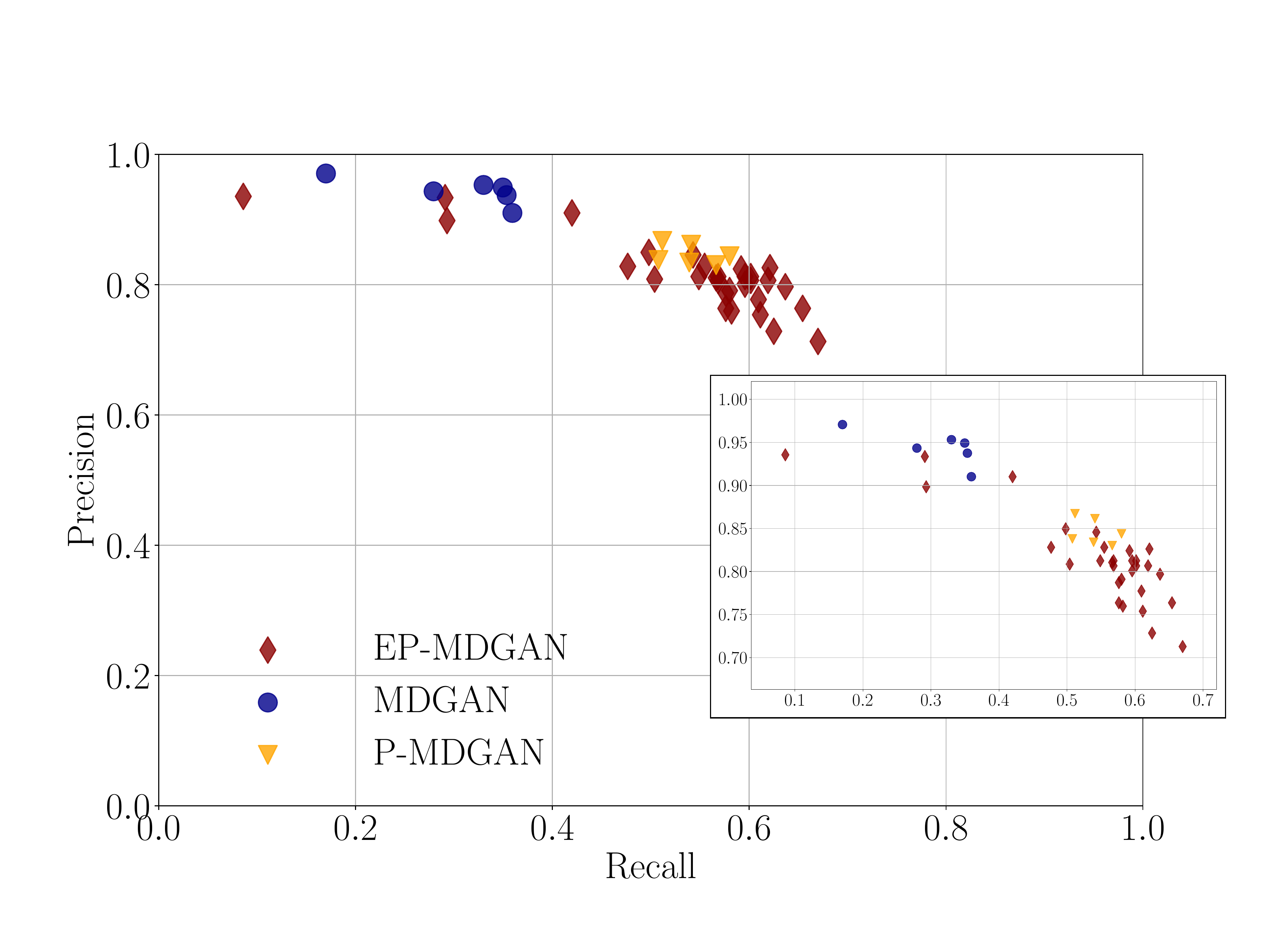}}%
		\end{tabular}
	\end{center}
	\caption{Precision and recall computed using the method proposed by \citep{PRD_improved} for the test set of the CIFAR10 (a) and F-MNIST (b) datasets.}%
	\label{fig:pr_scatter}
	\vskip -0.05in
\end{figure*}

\fboxrule=0pt
\begin{figure*}[t!]
	\begin{center}
		\begin{tabular}{cc}
			
			\subfigure[\rbigan]{\fbox{\includegraphics[width=0.5\linewidth]{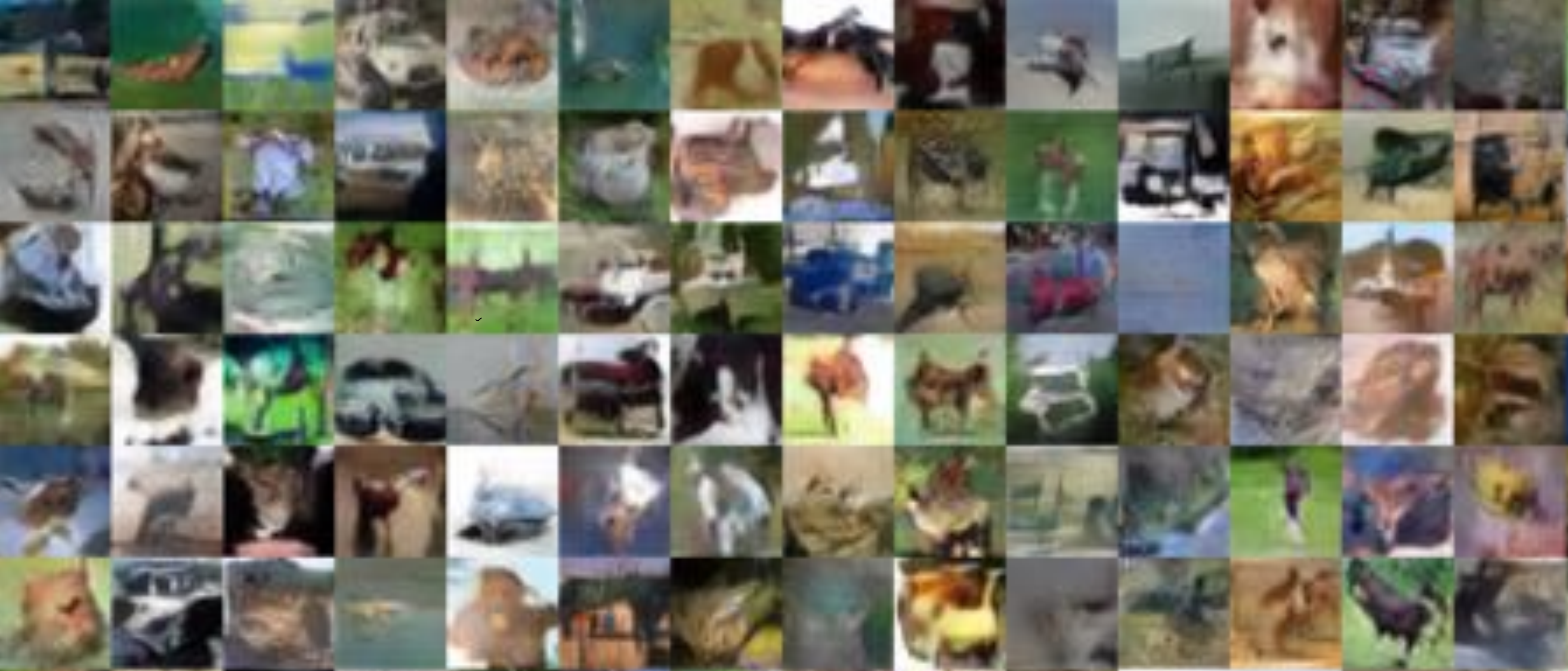}}}%
			\subfigure[\rbigan]{\fbox{\includegraphics[width=0.5\linewidth]{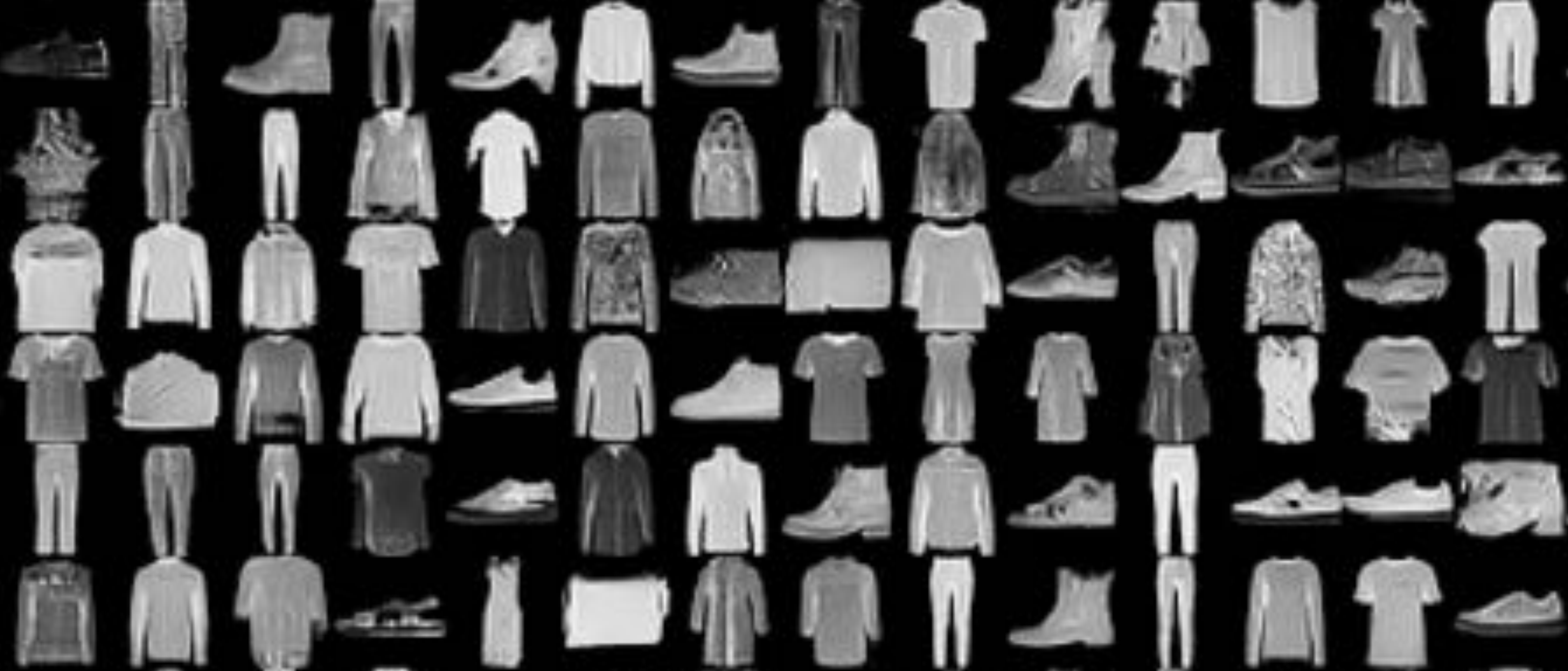}}}%
			\\
			\subfigure[\prbiganr]{\fbox{\includegraphics[width=0.5\linewidth]{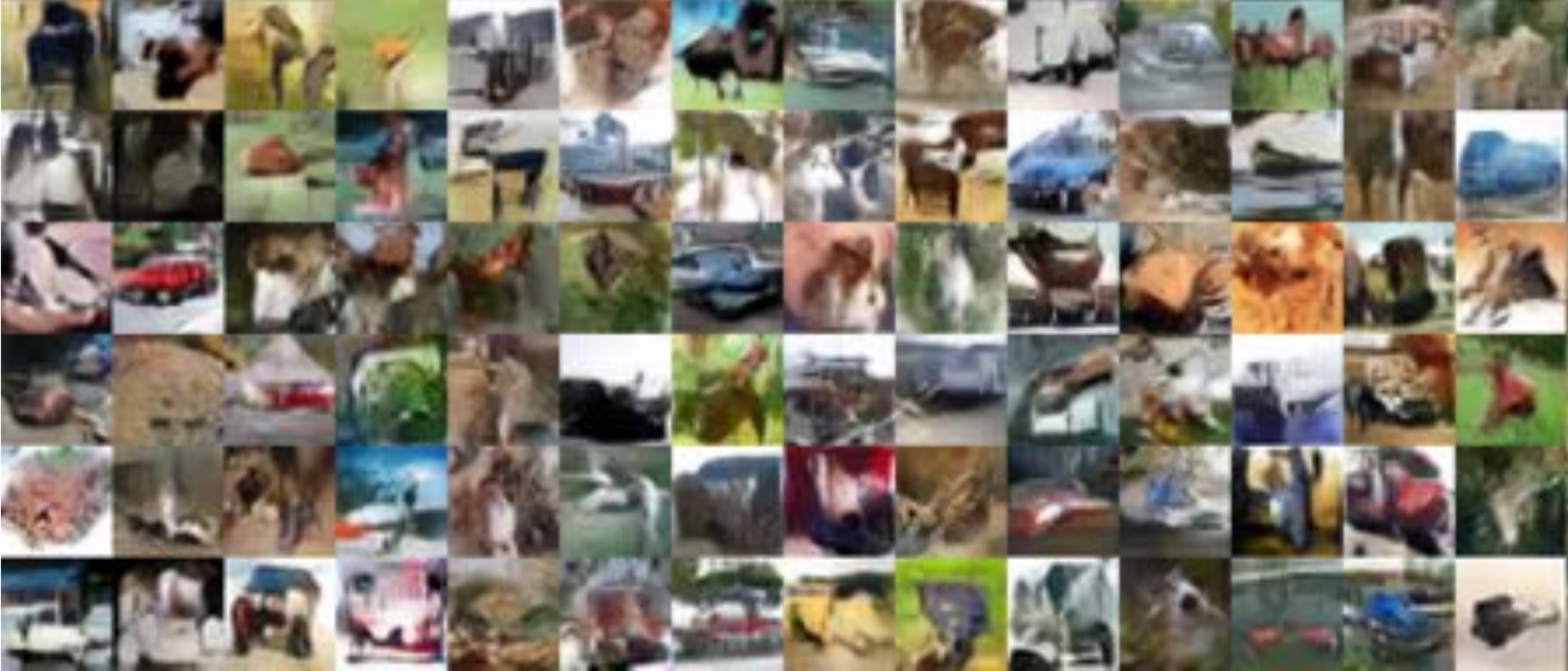}}}%
			\subfigure[\prbiganr]{\fbox{\includegraphics[width=0.5\linewidth]{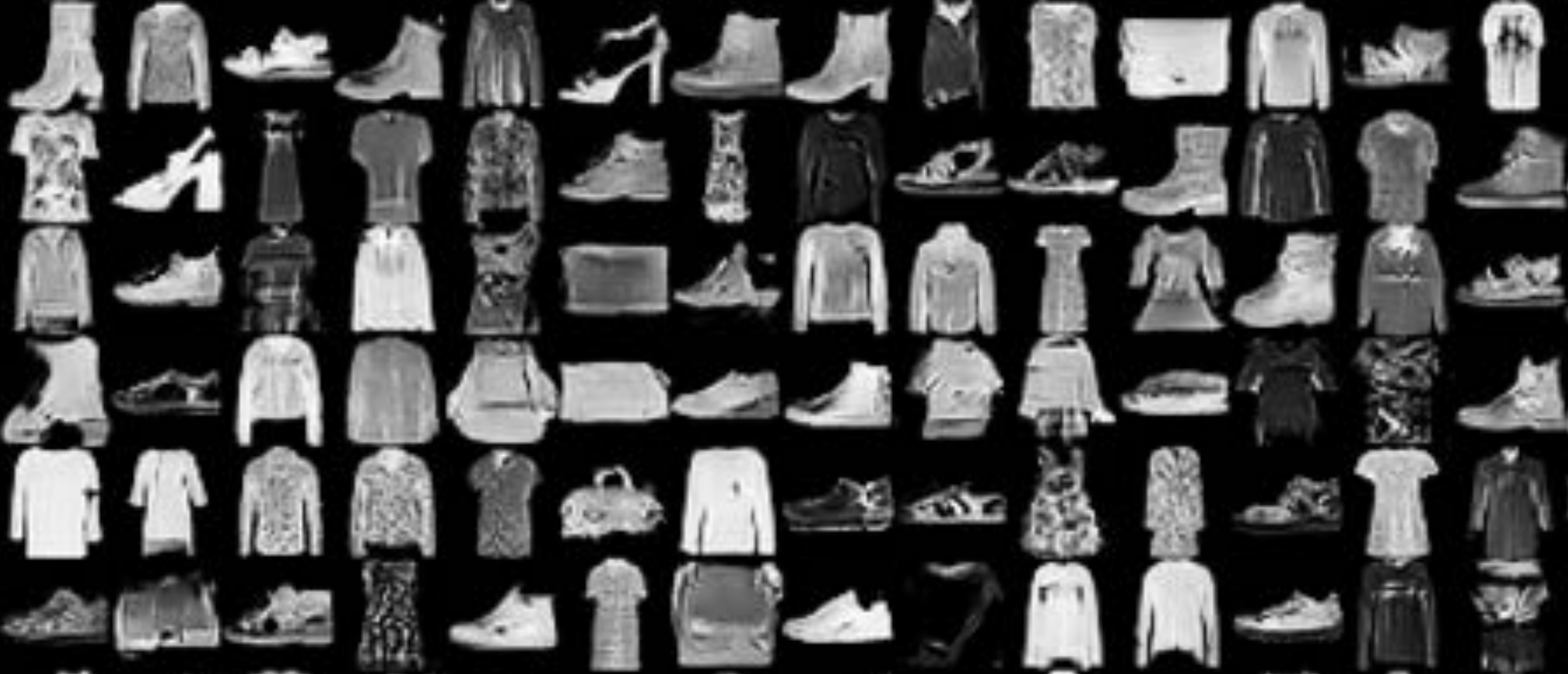}}}%
		\end{tabular}
	\end{center}
	\caption{Samples generated by \rbigan\ for CIFAR10 in (a) and F-MNIST (b), and \prbiganr\  for CIFAR10 in (c) and F-MNIST (d).}%
	\label{fig:x_gener}
	\vskip -0.05in
\end{figure*}

\begin{figure*}[t!]
	\begin{center}
		\begin{tabular}{cc}
			
			\subfigure[\rbigan]{\includegraphics[width=0.5\linewidth]{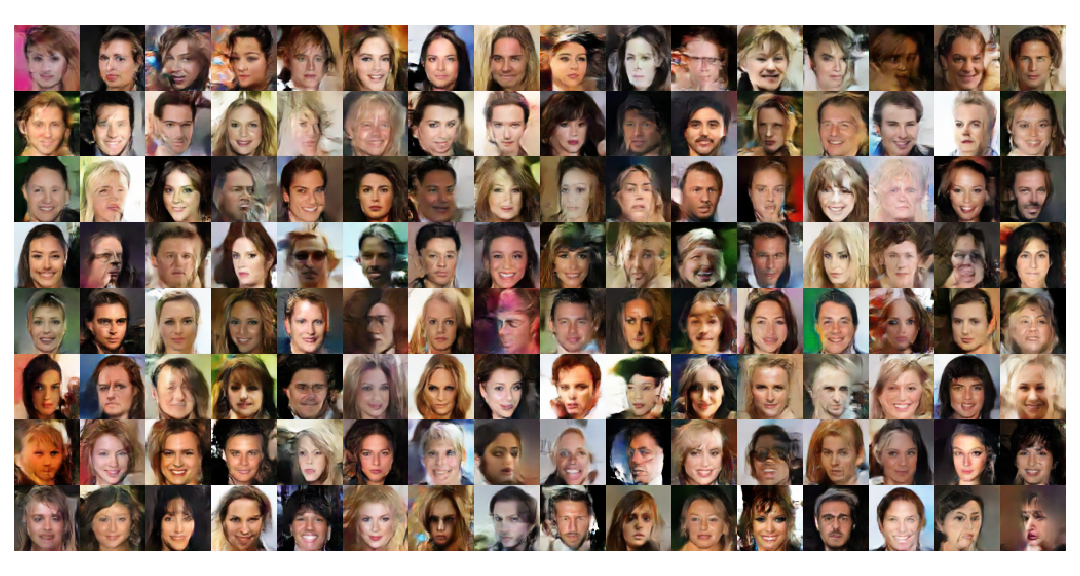}}%
			\subfigure[\prbiganr]{\includegraphics[width=0.5\linewidth]{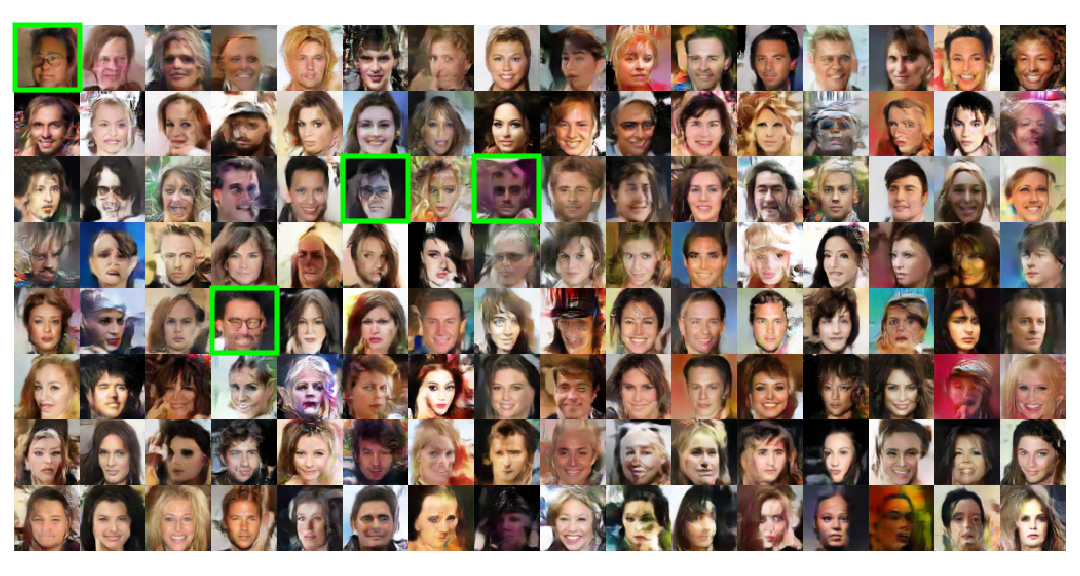}}%
		\end{tabular}
	\end{center}
	\caption{Samples generated by \rbigan\  (a) and \prbiganr\ (b) trained over the CelebA dataset. In (b), bordered in green images correspond to images of people wearing glasses, a feature not present even once in the 128 images generated by MDGAN. }%
	\label{fig:x_gener_cA}
	\vskip -0.05in
\end{figure*}

\subsection{Generated samples}
Figure  \ref{fig:fid_bar} shows the FID values computed over test samples of the CIFAR10 and F-MNIST datasets.  If we focus on  Figure  \ref{fig:fid_bar} (a), we can observe that the worst FID scores are obtained for \rbigan\ and P-MDGAN, and that the FID decreases (improves) when using any of the non-uniform sampling schemes, obtaining the best FID score for EP-MDGAN. Although these are the results for a specific configuration of the hyperpameters we have found that for most of the configurations these results hold. To gain more insights about the reason of this improvement, Figure \ref{fig:pr_scatter} shows a scatter plot of the Precision vs Recall scores obtained for all trained models. Observe the different \prbiganr\ models trained suffer a small precision drop and improved recall by over two-fold. This is an indication that the non-uniform sampling procedure encourages diversity in generated samples. The non-uniform sampling also provides a way to trade-off precision and recall in bi-directional GANs that can provide better diversity at the expense of limited image quality degradation.

We can also evaluate this improvement by visually inspecting generated samples, shown in Figure \ref{fig:x_gener} which shows samples generated for the CIFAR10 (left column) and F-MNIST (right column) datasets using the \rbigan\ (top row) and \prbiganr\ (bottom row). If we focus our attention to the left column, we observe samples generated using MDGAN, namely Figure \ref{fig:x_gener} (a), contain  homogeneous backgrounds  and simple objects with a spectrum of colors mainly limited to green, brown and blue tones. On the contrary, images produced by \prbiganr\ include diversity of color, and more complex backgrounds and shapes. This behavior is also acknowledged in the generated F-MNIST samples from \prbiganr, the images contain irregular patterns, textures, and more high heels. We also include in Figure \ref{fig:x_gener_cA} samples generated for the CelebA dataset. We can observe the \prbiganr\ generates faces with glasses (bordered in green) or other accessories, which similarly as with previous datasets can be considered as complex examples.

\begin{figure*}[t!]
	\begin{center}
		\begin{tabular}{cc}
			\subfigure[]{\includegraphics[width=0.48\linewidth]{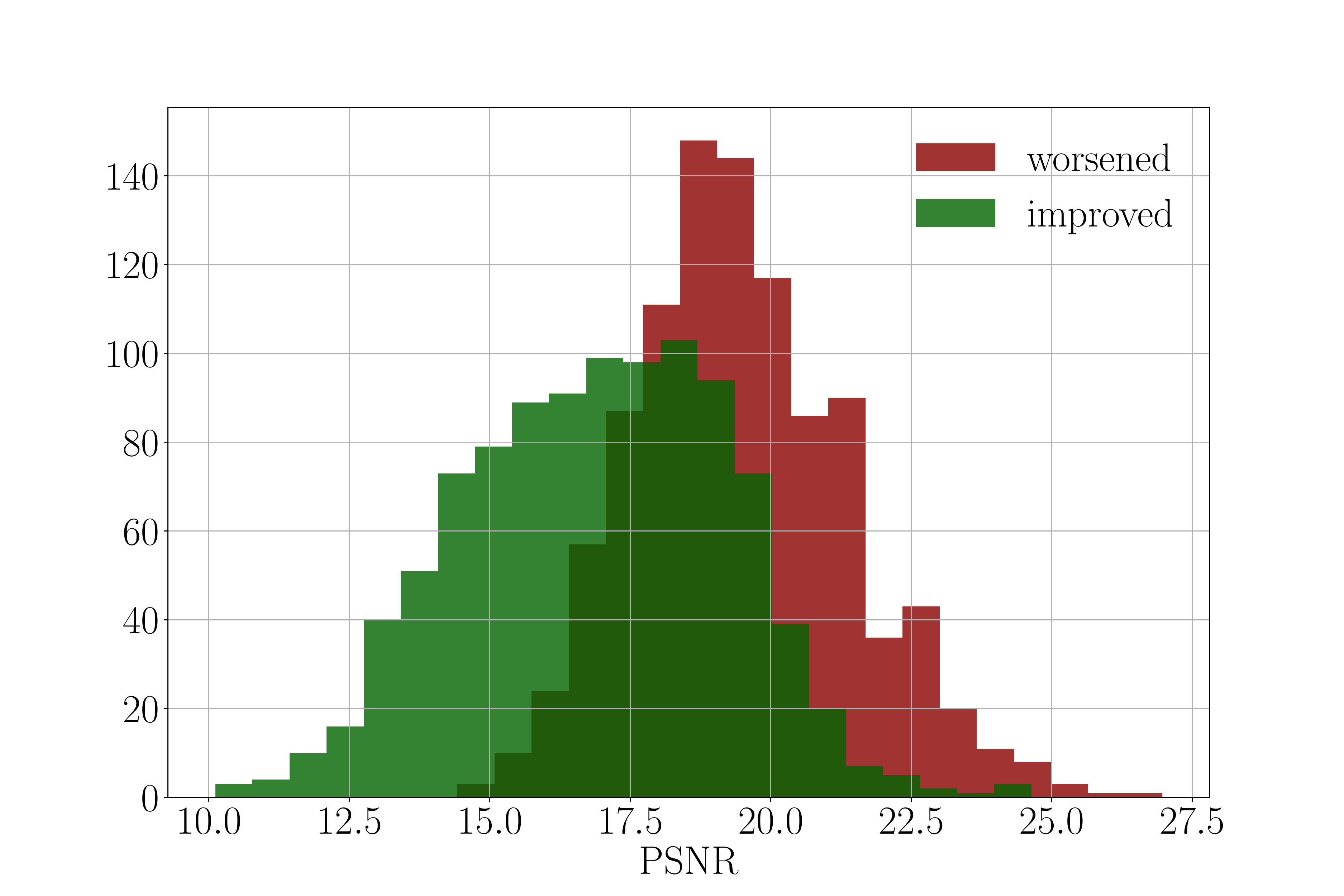}}%
			\subfigure[]{\includegraphics[width=0.48\linewidth]{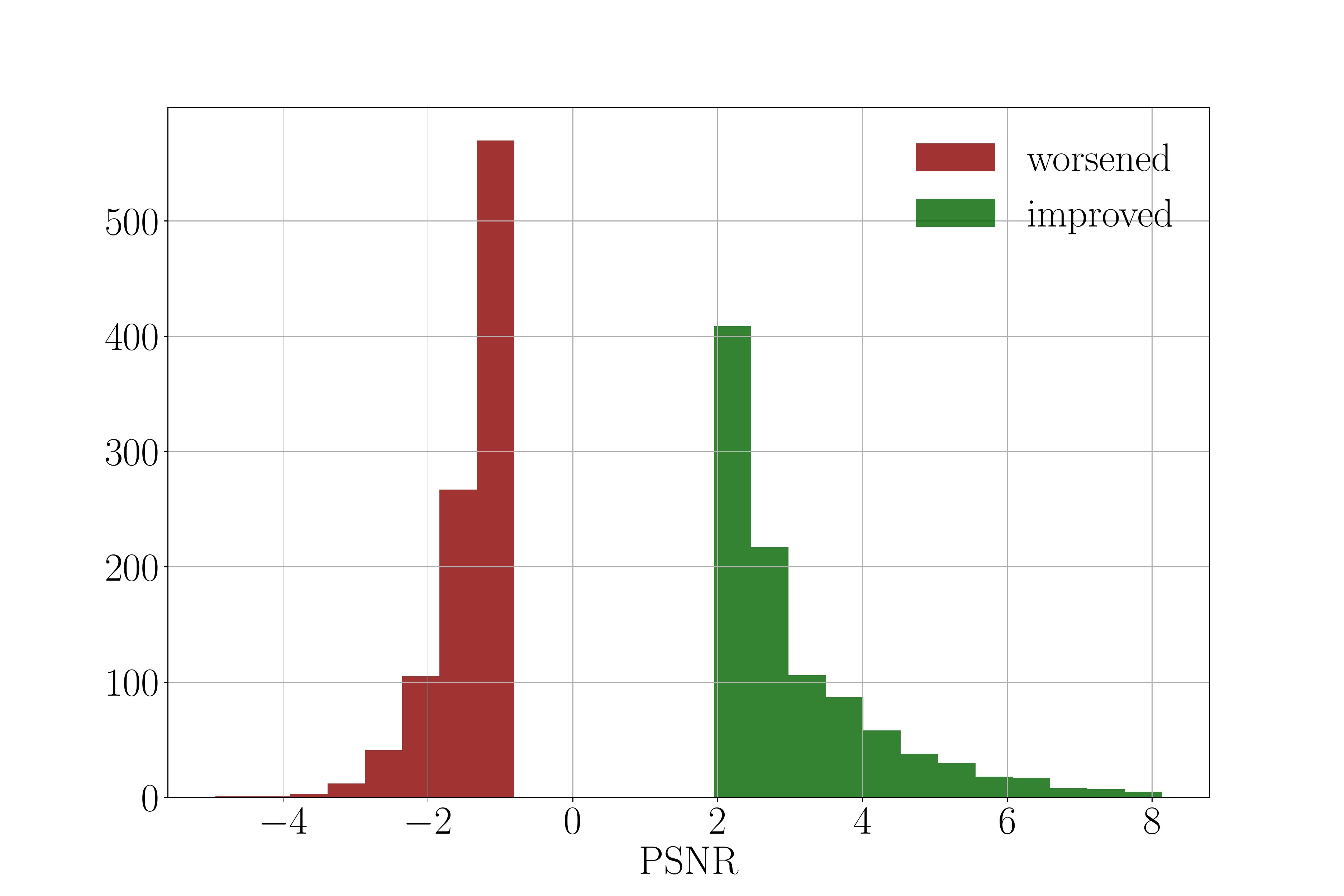}}%
		\end{tabular}
	\end{center}
	\caption{Two distributions of $\text{PSNR}(\x, G(E(\x)))$ between orignal samples and reconstructions obtained using \rbigan. In green (red) images that improve (decrease) the PSNR when using the reconstruction from \prbiganr. The distribution in green (red) shows the  difference in $\text{PSNR}(\x, G(E(\x)))$ using the reconstructions obtained with \prbiganr\ and \rbigan\ for the $10\%$ of images that improve (worsen) the most the $\text{PSNR}$ when using \prbiganr.}%
	\label{fig:hist_psnr_impr}
\end{figure*}

\begin{figure}[t!]
	\centering
	\subfigure[CIFAR10]{\includegraphics[width=0.35\linewidth]{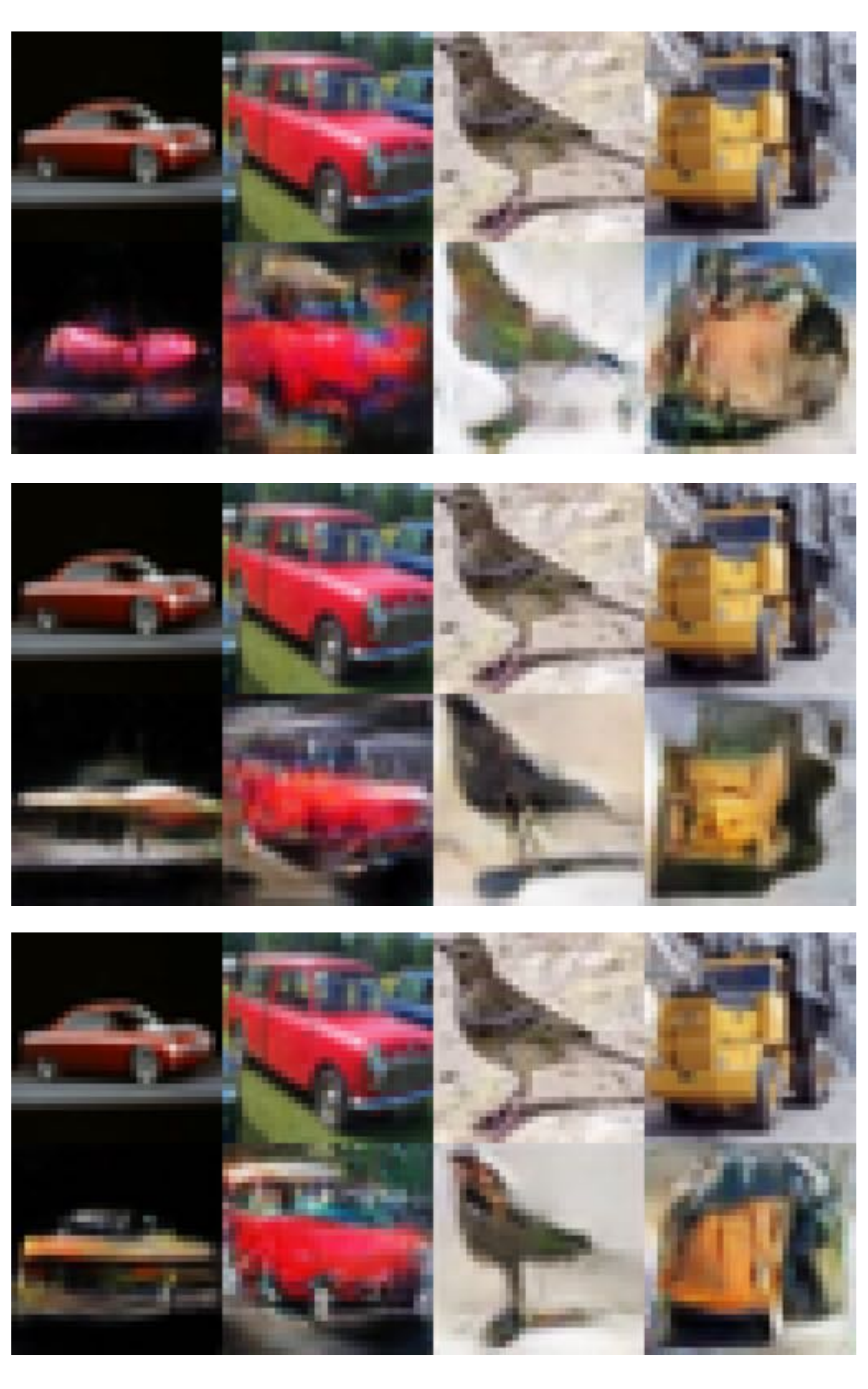}}%
	\hspace{2.5cm}
	\subfigure[F-MNIST]{\includegraphics[width=0.35\linewidth]{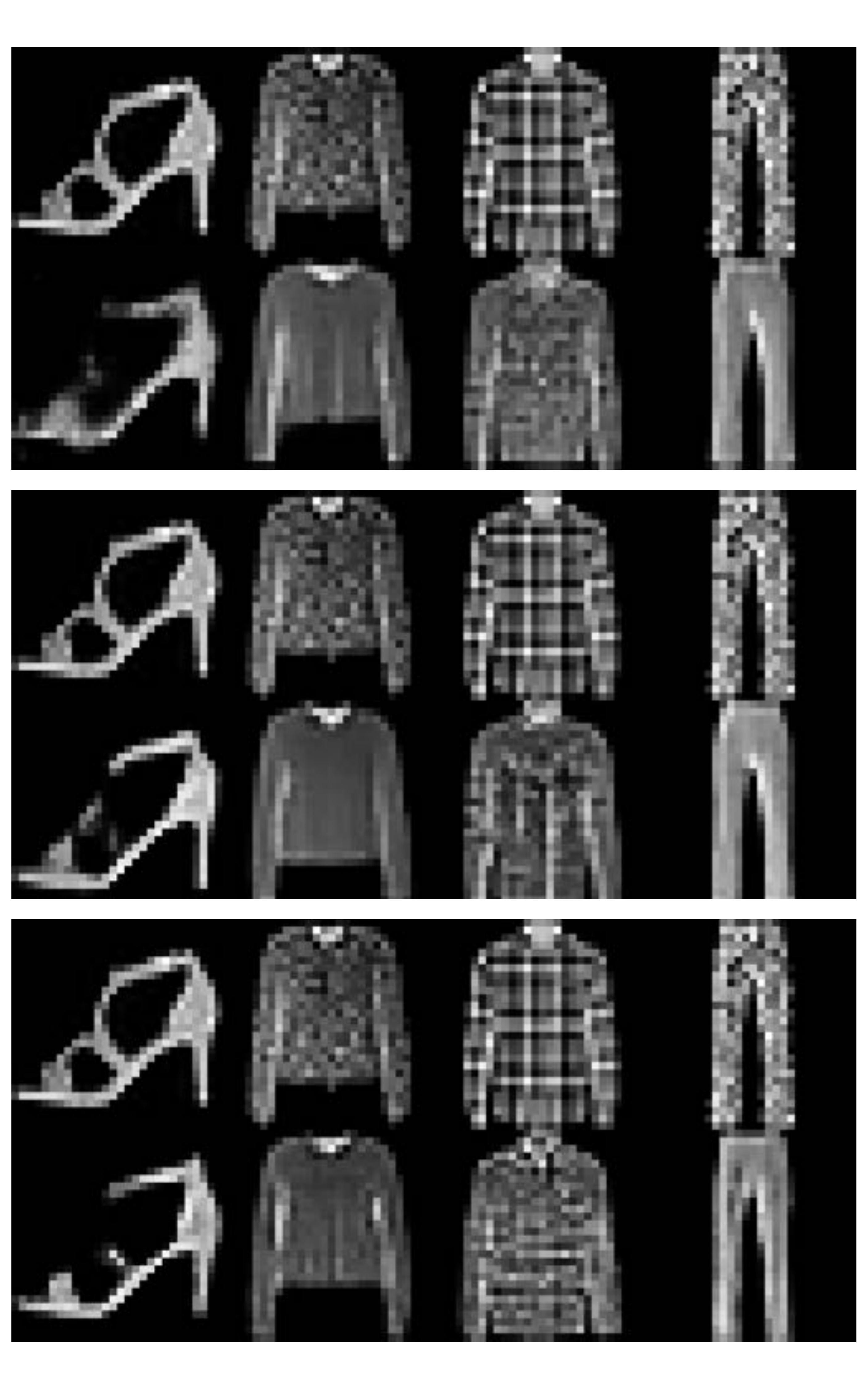}}%
	
	\caption{Samples from the test set $\x$ along with their corresponding reconstruction  $G(E(\x))$ from the Cifar10 (a) and F-MNIST (b) datasets.  From top to bottom: \rbigan, \prbigan\ and \prbiganr.}%
	\label{fig:x_recons}
	\vskip -0.05in
\end{figure}

\textbf{Reconstruction quality:}
 Figure \ref{fig:hist_psnr_impr} (a) shows two histograms of the PSNR computed between  CIFAR10 test images and their corresponding MDGAN reconstruction. The histogram in green represents the $10\%$ of images that improve the reconstruction PSNR using \prbiganr\ the most. Similarly, the red histogram represents the $10\%$ of images that worsen the most the reconstruction PSNR using \prbiganr. 

 These results shows that the non-uniform sampling training encourages better reconstruction of the images that obtained worse reconstruction quality, i.e. visually complex examples, when using the baseline. As such, a slight decrease of reconstruction quality in the subset of simpler images does not bring a noticeable visual impact. Figure \ref{fig:hist_psnr_impr} (b) shows the order or improvement/deterioration. 

 In Figure \ref{fig:x_recons} we show 4 visually complex images from the CIFAR10 and F-MNIST test sets and their corresponding reconstructions. We observe that \prbigan\ and \prbiganr\ achieve better qualitative reconstruction quality than \rbigan.

\section{Conclusion}
\label{sec:discussion}

In this paper we have first proposed a novel regularization on the inference network of the BiGAN so that the distribution of the encoded latent space variables matches the prior distribution. This means that the latent representation of samples lays in the typical set and the corresponding reconstruction is likely to be sampled from the generative model. We called the resulting model \prbigan.

Then we introduce a method to estimate marginal log-likelihood over reconstructed images, rather than original samples. We provide an explanation of why log-likelihood and reconstruction quality should be computed independently so that the latter does not bias the first. The reproduction quality tells us if a sample can be generated by the GAN and how good it matches the test sample\footnote{This measure had been proposed previously in \cite{zhu2016, Metz17}, but has not been advocated for systematically evaluating GANs.}. The estimation of the log-likelihood of the reconstruction tells us how likely are we to see that reconstruction, which is the only image that the GAN can produce. Estimating the likelihood of the test sample directly is much harder and it mixes these two relevant metrics in one, making it useless to evaluate GANs, as already point it out in \cite{EGM1}.
 
The results in log-likelihood estimation show that training and test samples suffer significant over and under-representation issues that need to be corrected when training GANs. We have also noticed that the samples that are more visually complex lead to higher reconstruction error and lower marginal likelihoods. For example, we can argue that the samples that present lower marginal likelihood can be over-sampled when training GANs, as we should not expect that harder to generate samples need to be seen an equal number of times that those that are easier to generate.  We finally proposed to train the GANs using a non-uniform sampling scheme and we propose two approaches for that, one of them static (for computational reasons)  based on the estimated log-likelihoods, and the other one dynamical using the PSNR between original images and reconstructions. We show they improve the baseline in terms of image diversity while also improving the reconstruction of visually complex images.

% Acknowledgements should go at the end, before appendices and references  
\acks{
	The work of Pablo M. Olmos is supported by Spanish government MCI under grant PID2019-108539RB-C22 and RTI2018-099655-B-100, by Comunidad de Madrid under grants IND2017/TIC-7618, IND2018/TIC-9649, and Y2018/TCS-4705, by BBVA Foundation under the Deep-DARWiN project, and by the European Union (FEDER and the European Research Council (ERC) through the European Unions Horizon 2020 research and innovation program under Grant 714161). We also gratefully acknowledge the support of NVIDIA Corporation with the donation of the Titan X Pascal GPU used for this research.
 }

\vskip 0.2in
\bibliography{bibliography}

\appendix
\newpage

\section{Experiment details}
\label{app:architecture}

In this section we provide further information about the arquitecture of the networks and training specifications used in our experiments.

\subsection*{Architecture of GANs}

The three main blocks, $G(\cdot)$, $D(\cdot)$ and $E(\cdot)$, are convolutional neural networks. Considering the networks for a 32x32 image as the reference, the networks for a 28x28 image only change in the second layer of the generator and the networks for a 64x64 image only differ in one layer each. The complete details are in Tables \ref{table:G32_architecture}, \ref{table:D_architecture} and \ref{table:E_architecture}. Following the same approach proposed in the original BiGAN paper, $D(\z, \x)$ takes $\x$ as input, after some convolutional layers $\z$ is concatenated, and the resulted matrix is passed through several convolutional layers. We use LeakyReLU as activation function with a slope of 0.1 and Spectral normalization. The generator is a 5 layer deconvolutional layer with batch normalization and ReLU as activation function. The inference network $E(\x)$ is a 7 layer convolutional network with Leaky ReLU as activation function, followed by a dense layer.

\begin{table}[h!]
	\centering
	\begin{tabular}{c} 
		
		\hline\hline
		$\z \in  \mathbb{R}^{256 \times 1 \times  1} \sim  N(\boldmath{0}, \boldmath{ I })$ \\
		\hline
		deconv, stride=1, pad=0, $4 $, 512, BN, ReLU\\
		\hline
		deconv, stride=2, pad=1, $4$ [$3$] , 256, BN, ReLU\\
		\hline
		deconv, stride=2, pad=1, $4 $, 128, BN, ReLU\\
		\hline
		deconv, stride=2, pad=1, $4$, 64, BN, ReLU\\
		\hline
		deconv, stride=1 (2) , pad=1, $3 $ ($4$), 3, Tanh\\
		\hline\hline
	\end{tabular}
	\caption{Generator 32 (64)[28] }
	\label{table:G32_architecture}
\end{table}

\begin{table}[h!]
	\centering
	\begin{tabular}{c} 		
		\hline\hline
		$\x \in  \mathbb{R}^{3 \times H \times  W} $ \\
		\hline
		conv, stride=1, pad=1, $3 $, 64, SN, LReLU(0.1)\\
		\hline
		conv, stride=2, pad=1, $4 $, 64, SN, LReLU(0.1)\\
		\hline
		conv, stride=1 (2), pad=1, $4 $, 128, SN, LReLU(0.1)\\
		\hline
		conv, stride=2, pad=1, $4 $, 128, SN, LReLU(0.1)\\
		\hline
		conv, stride=1, pad=0, $4 $, 256, SN, LReLU(0.1)\\
		\hline
		conv, stride=2, pad=0, $4 $, 256, SN, LReLU(0.1)\\
		\hline	\hline
		$\z \in  \mathbb{R}^{256 \times 1 \times  1} \sim  N(\boldmath{0}, \boldmath{ I })$ \\
		\hline
		conv, stride=1, pad=0, $1 $, 256, SN, LReLU(0.1)\\
		\hline	\hline
		conv, stride=1, pad=0, $1 $, 512, SN, LReLU(0.1)\\
		\hline
		conv, stride=1, pad=0, $1 $, 1024, SN, LReLU(0.1)\\
		\hline
		dense,  $1024 \times 1$\\
		\hline\hline
	\end{tabular}
	\caption{Discriminator 32 (64)}
	\label{table:D_architecture}
\end{table}

\begin{table}[h!]
	\centering
	\begin{tabular}{c} 
		
		\hline\hline
		$\x \in  \mathbb{R}^{3 \times H \times  W} $ \\
		\hline
		conv, stride=1, pad=1, $3 $, 64,  LReLU(0.1)\\
		\hline
		conv, stride=2, pad=1, $4 $, 64,  LReLU(0.1)\\
		\hline
		conv, stride=1, pad=1, $3 $, 128,  LReLU(0.1)\\
		\hline
		conv, stride=2, pad=1, $4$, 128,  LReLU(0.1)\\
		\hline
		conv, stride=1 (2), pad=1, $3 $, 256,  LReLU(0.1)\\
		\hline
		conv, stride=2 , pad=1, $4 $, 256,  LReLU(0.1)\\
		\hline
		conv, stride=1 , pad=1, $3 $, 512,  LReLU(0.1)\\	
		\hline
		dense,  $4*4*512 \times 256$\\	
		\hline\hline
	\end{tabular}
	\caption{Inference 32 (64)}
	\label{table:E_architecture}
\end{table}

\section{Training details}
\label{app:training_details}
The  objective function for the PR-BiGAN is 
\begin{equation}
\begin{split}
\label{eq:bigan_obj}
\Lm ( D , G, E ) &=  \Lm_{ \mathrm {BiGAN} } +   \lcyc \mathbb { E } _ { \x \sim p _ { r } ( \x ) } [ d(G(E(\x)), \x)] \\
& +\lno \mathbb { E } _ { \x \sim p _ { r } ( \x ) } \left[  \left(|| E(\x)||_2-\sqrt{\text{dim}(\z)} \right)^2 \right]
\end{split}
\end{equation}

where the MSE is used as the distance metric in the reconstruction loss. If we set $\lno=0$ we obtain the \rbigan and if we set $\lcyc=0$ the resulting model is the BiGAN. Additionally, we add $\lperc$ and $\ldist$ to perform non-uniform sampling. Basically, these  hyperparameters control the  different regularization techniques. The parameter $\lcyc \in\mathbb{R}_{\geq0}$ controls the reconstruction loss; $\lno \in \mathbb{R}_{\geq0}$ adapts the regularization on the norm of  the distribution induced on the encoder $||E(\x)||$  so that it matches the prior distribution; the parameter  $\lperc \in [0,1]$ fixes the percentage of uniform samples used in each batch during training; and $\ldist\geq0$ determines the distribution from which to draw non-uniform samples. This distribution has been computed using the estimated log-likelihoods $ LL(\x_i)$ for a given data set. Firstly we order the images from largest loglikelihood to smaller, $LL(\x_1) > ... >LL(\x_k)> ....> LL(\x_N) $, and we keep the location in this list $k$. Then we use this index $k$ to compute their probability of being sampled: $\ldist\geq0$. Additionally, we introduce a second hyperparameter, $\lperc\in[0,1]$, to control the percentage of samples per mini-batch that use this re-weighting strategy, while the remaining are chosen uniformly.

\begin{align}
\label{eq:non_uniform}
Pr(\x_k | LL(G(E(\x_k))), \ldist, \lperc)= \frac{1-\lperc}{N}+\lperc\frac{\left(k/100 \right)^{\ldist}}{\sum_{j=1}^{N} \left(j/100 \right)^{\ldist}}.\notag
\end{align}

We have explored different configurations of the regularization hyperparameters $\lcyc$, $\lperc$, $\ldist$ in order to demonstrate its performance. They are summarized in Table \ref{table:hyperparameters}. The regularization paramter $\lno$ is automatically learnt during training in order to motivate the distribution obtained from $E(\x)$ matchs the noise distribution $p(\z) = N(\z | \boldmath{0}, \boldmath{ I })$. It is initialize with a small value $\lno= 0.01$ which is kept constant for the first 200 epochs and then it is updated according to the variance of $||E(\x)||$ computer over all the examples from the train set compared to the variance of  $||\z||$. For all the experiments, we fix the \textit{learning rate } $2e^{-4}$, the \textit{batch size} 128, $\dim(\z)=256$. We train the models for 800 \textit{epochs}.

\begin{table}[!htb]
	\begin{minipage}{.5\linewidth}
		\centering
\begin{tabular}{|c|c|c|c|c|c|}
	\multicolumn{2}{c }{}&  \multicolumn{4}{|c| }{$\lambda_{\text{dist}}$}\\
	\hline
	$\lambda_{\text{cyc}}$&$\lambda_{\text{perc}}$ & 4  & 8  &12 & 16\\
	\hline
	& 0.2&\checkmark  &&& \checkmark\\
	\cline{2-6}
	3 & 0.5&  &&\checkmark&\\	
	\cline{2-6}
	& 0.8& \checkmark &  & \checkmark & \\
	\hline
	\hline
	& 0.2& \checkmark  &\checkmark&&\\
	\cline{2-6}
	5 & 0.5&  & \checkmark&\checkmark&\\	
	\cline{2-6}
	& 0.8&  &&&\checkmark\\
	\hline
	\hline
	& 0.2&  &&\checkmark&\checkmark\\
	\cline{2-6}
	6 & 0.5&\checkmark  &&&\\	
	\cline{2-6}
	& 0.8&  &&\checkmark&\checkmark\\
	\hline
	\hline
	& 0.2&  &&&\checkmark\\
	\cline{2-6}
	7 & 0.5&  &&\checkmark&\checkmark\\	
	\cline{2-6}
	& 0.8&  &&\checkmark&\checkmark\\
	\hline
	\hline
	& 0.2&  &\checkmark&&\\
	\cline{2-6}
	8 & 0.5&\checkmark  &&\checkmark&\\	
	\cline{2-6}
	& 0.8&  &&\checkmark&\checkmark\\
	\hline
	\hline
	& 0.2&  &\checkmark&\checkmark&\\
	\cline{2-6}
	9 & 0.5&  & &\checkmark& \\	
	\cline{2-6}
	& 0.8&  &\checkmark&& \checkmark \\
	\hline
\end{tabular}
\caption*{(a) Cifar10}
\label{table:hyperparameters_c10}
	\end{minipage}%
\begin{minipage}{.5\linewidth}
	\centering
	\begin{tabular}{|c|c|c|c|c|c|}
	\multicolumn{2}{c }{}&  \multicolumn{4}{|c| }{$\lambda_{\text{dist}}$}\\
	\hline
	$\lambda_{\text{cyc}}$&$\lambda_{\text{perc}}$ & 4  & 8  &12 & 16\\
	\hline
	& 0.2&  &&&\checkmark \\
	\cline{2-6}
	3 & 0.5& \checkmark &\checkmark&&\\	
	\cline{2-6}
	& 0.8&\checkmark  &\checkmark&& \\
	\hline
	\hline
	& 0.2&  &&\checkmark &\\
	\cline{2-6}
	5 & 0.5&  &&\checkmark&\checkmark\\	
	\cline{2-6}
	& 0.8&  &&\checkmark&\checkmark\\
	\hline
	\hline
	& 0.2&  &\checkmark&\checkmark &\checkmark\\
	\cline{2-6}
	6 & 0.5&  &&&\\	
	\cline{2-6}
	& 0.8&  & &\checkmark&\checkmark\\
	\hline
	\hline
	& 0.2&  &&&\checkmark\\
	\cline{2-6}
	7 & 0.5&  &\checkmark&\checkmark&\\	
	\cline{2-6}
	& 0.8& \checkmark &\checkmark&&\\
	\hline
	\hline
	& 0.2&  &&&\\
	\cline{2-6}
	8 & 0.5&\checkmark  &\checkmark&&\checkmark\\	
	\cline{2-6}
	& 0.8&  &&\checkmark&\checkmark\\
	\hline
	\hline
	& 0.2&  &&&\checkmark\\
	\cline{2-6}
	9 & 0.5&  &\checkmark&\checkmark&\checkmark\\	
	\cline{2-6}
	& 0.8&  &&\checkmark&\\
	\hline
\end{tabular}
\caption*{(b) F-MNIST}
\label{table:hyperparameters_fm}
	\end{minipage} 
\\
\centering
\begin{minipage}{.5\linewidth}
	\centering
		\begin{tabular}{|c|c|c|c|c|c|}
		\multicolumn{2}{c }{}& \multicolumn{4}{|c| }{$\lambda_{\text{dist}}$}\\
		\hline
		$\lambda_{\text{cyc}}$&$\lambda_{\text{perc}}$ & 4  & 8  &12 & 16\\
		\hline
		& 0.2& &\checkmark&&\checkmark\\
		\cline{2-6}
		3 & 0.5&\checkmark &&&\checkmark\\	
		\cline{2-6}
		& 0.8&\checkmark &&&\\
		\hline
		\hline
		& 0.2&  &\checkmark&&\checkmark\\
		\cline{2-6}
		5 & 0.5&  &&&\checkmark\\	
		\cline{2-6}
		& 0.8&  &\checkmark&&\checkmark\\
		\hline
		\hline
		& 0.2&  &\checkmark&&\checkmark\\
		\cline{2-6}
		7 & 0.5&  &&\checkmark&\\	
		\cline{2-6}
		& 0.8& \checkmark &\checkmark&&\\
		\hline
	\end{tabular}
	\caption*{(c) CelebA}
	\label{table:hyperparameters_cA}
\end{minipage}%

	\caption{Set of hyperparameters validated}
\label{table:hyperparameters}
\end{table}

We use Adam optimizer with parameters $\beta_1 = 0.5$ and $\beta_2=0.999$ and Exponential Learning rate decay 0.99 starting at epoch 400. We alternate between the optimization of the $D(\cdot)$ parameters and the optimization of the $E(\cdot)$ and $G(\cdot)$ parameters jointly. We perform 5 updates of $D(\cdot)$  per 1 update of $G(\cdot)$ and $E(\cdot)$.

\section{Additional Results}
\label{app:results}
\begin{figure*}[t!]
	\begin{center}
		\begin{tabular}{cc}
			
			\subfigure[Test set]{\includegraphics[width=0.5\linewidth]{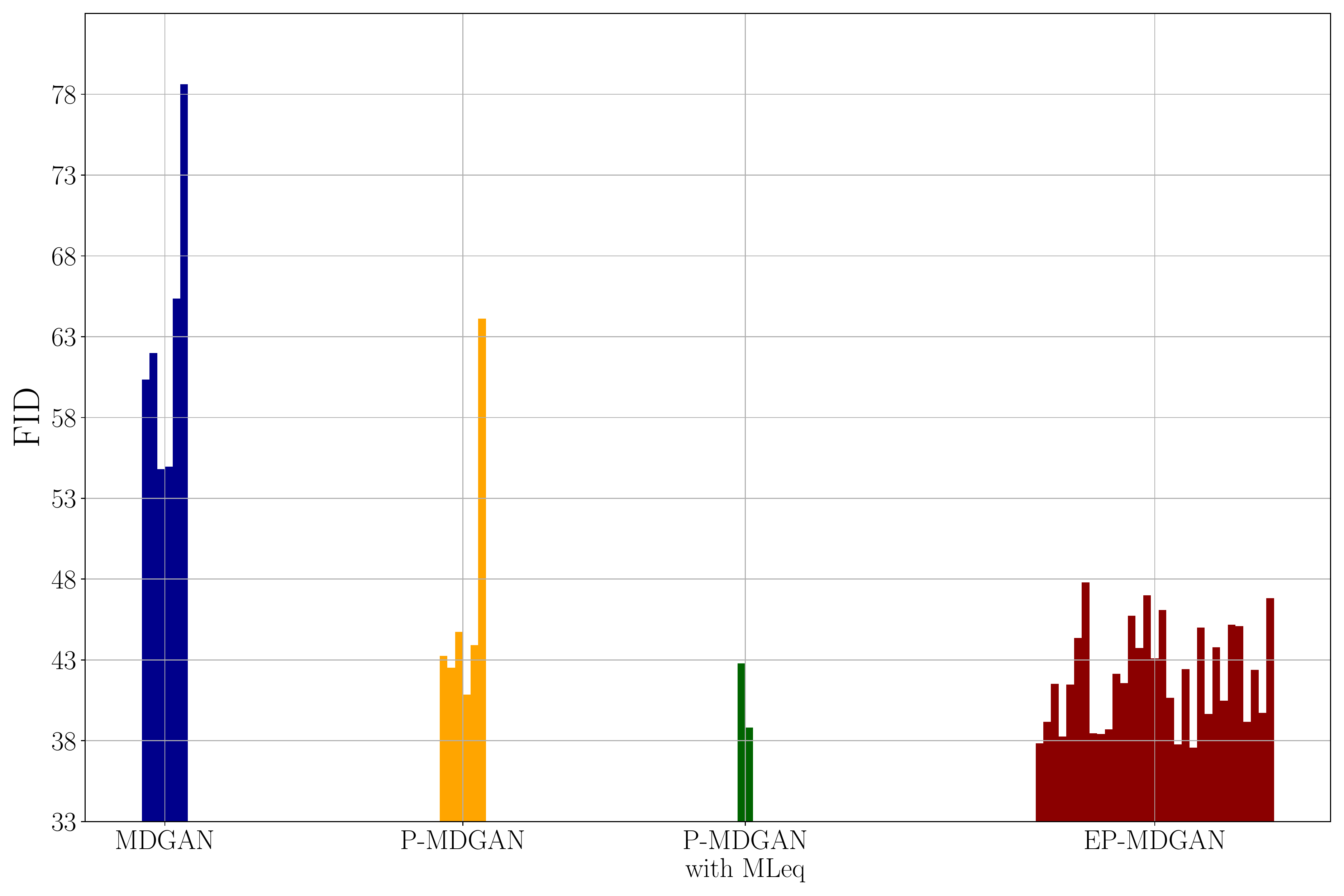}}%
			\subfigure[Train set]{\includegraphics[width=0.5\linewidth]{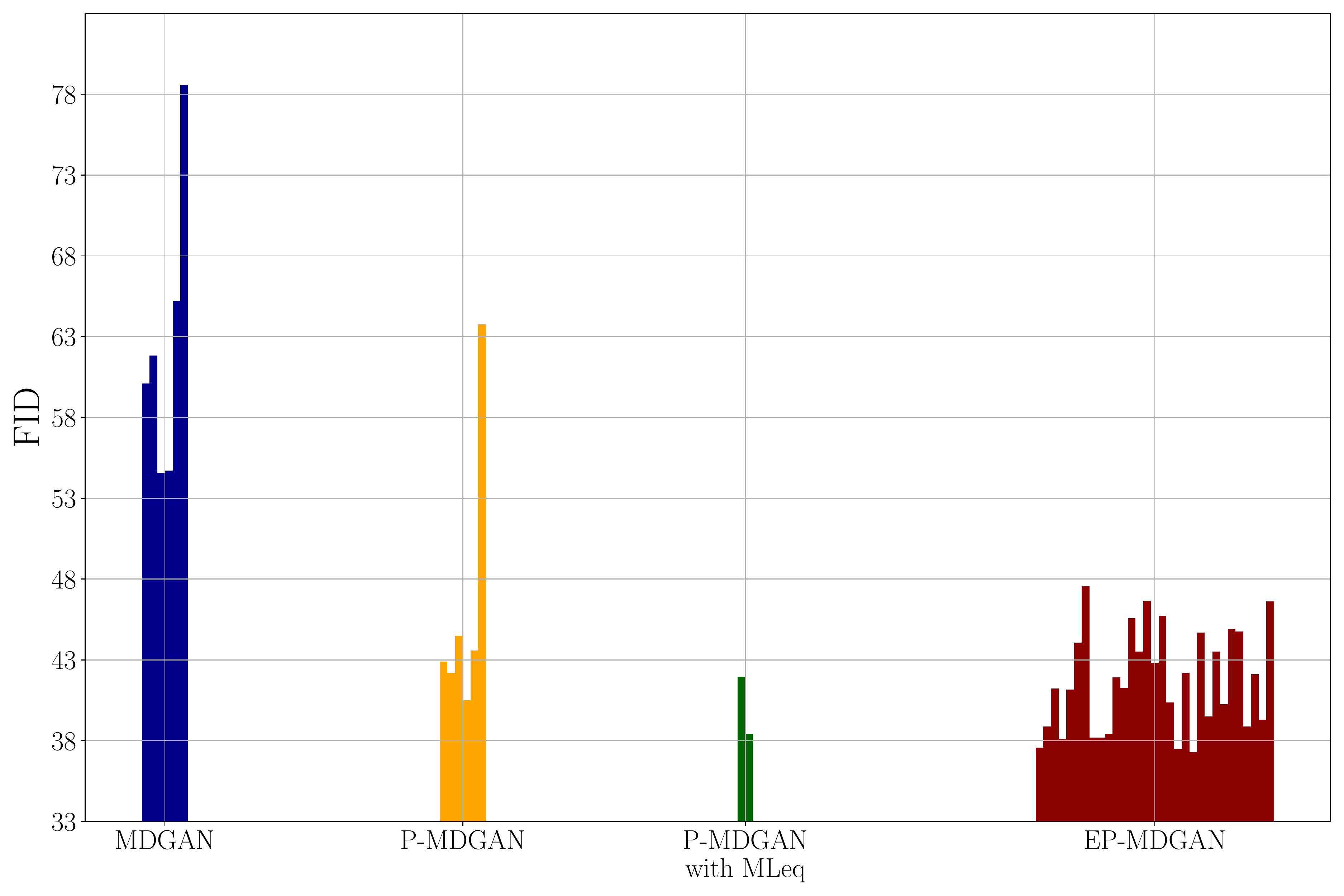}}%
		\end{tabular}
	\end{center}
	\caption{FID scores computed using the test set in (a)  and the train set in (b) of the CIFAR10 dataset. }%
\label{fig:fid_scores}
	\vskip -0.05in
\end{figure*}

In this section we extend the results presented in the main paper. For all the models trained with the CIFAR10 dataset, we show the FID scores obtained using the test set in Figure \ref{fig:fid_scores} (a) and the train set in  Figure \ref{fig:fid_scores} (b). Observe that all the configurations of \prbiganr\ trained obtain low and similar FID values while \rbigan\ and \prbigan\ suffer from high variance, the choice of the $\lcyc$ affects the FID scores.

\begin{figure*}[h!]
	\begin{center}
		\begin{tabular}{cc}
			
			\subfigure[Orignal]{\includegraphics[width=0.5\linewidth]{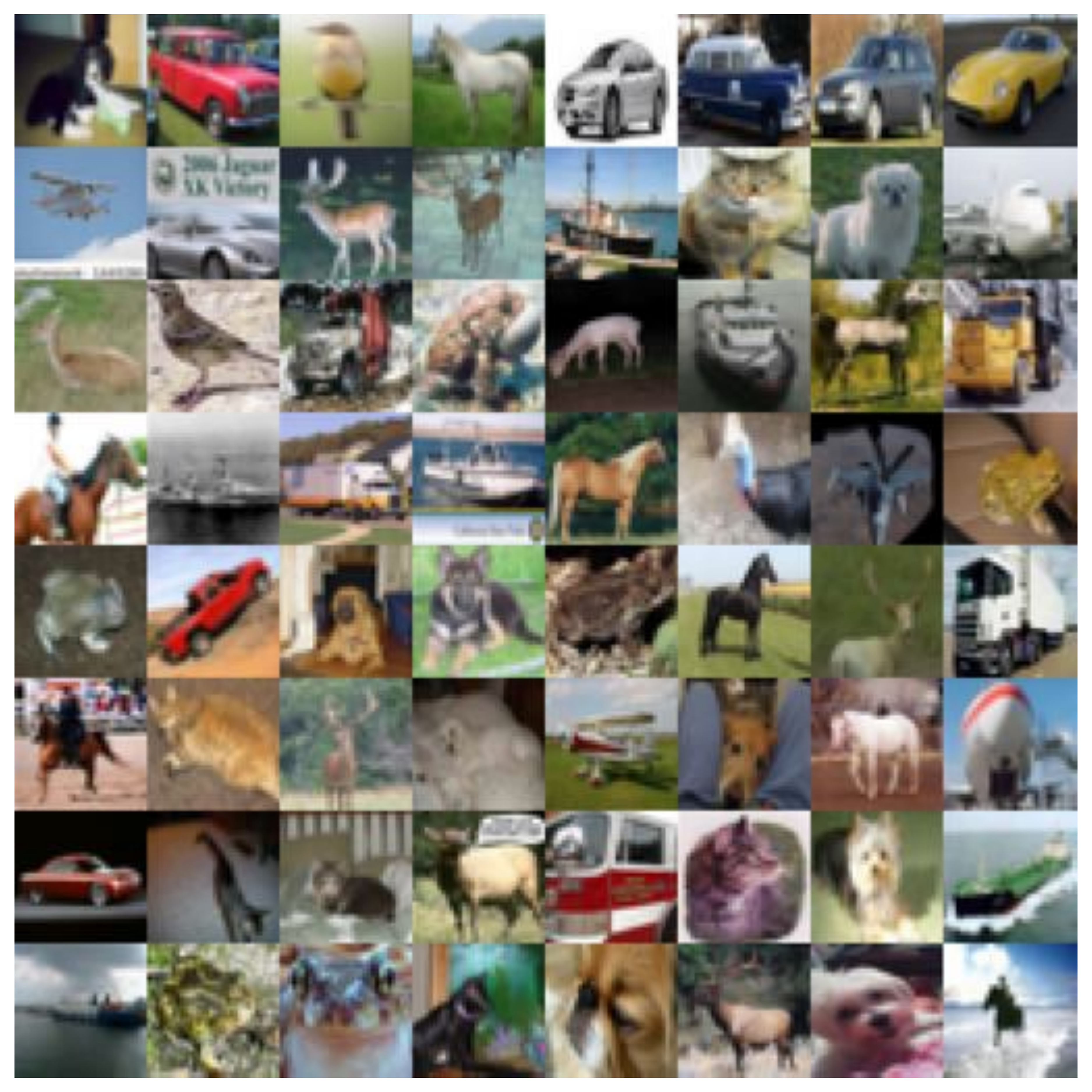}}%
			\subfigure[\prbiganr]{\includegraphics[width=0.5\linewidth]{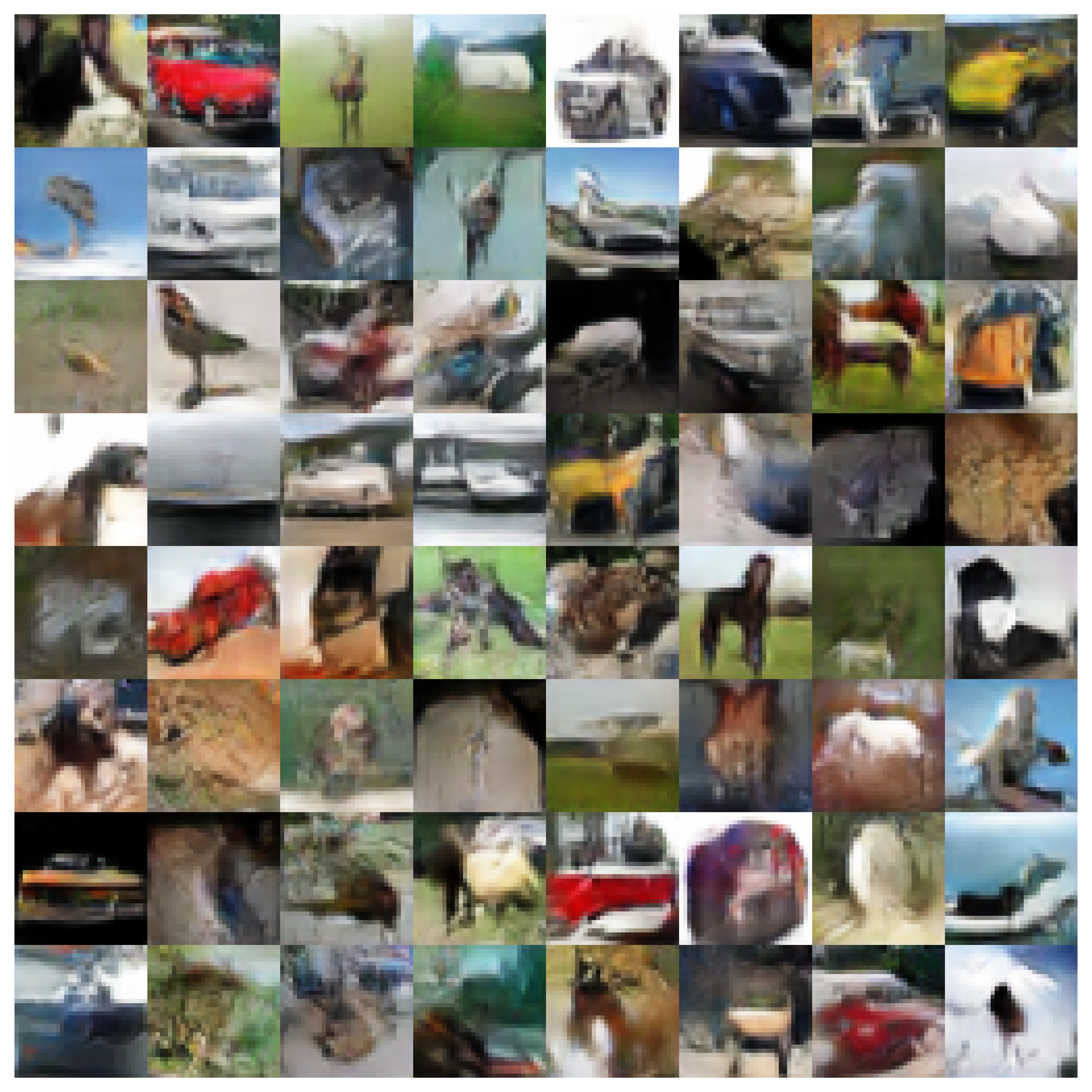}}%
			\\
			\subfigure[\prbigan]{\includegraphics[width=0.5\linewidth]{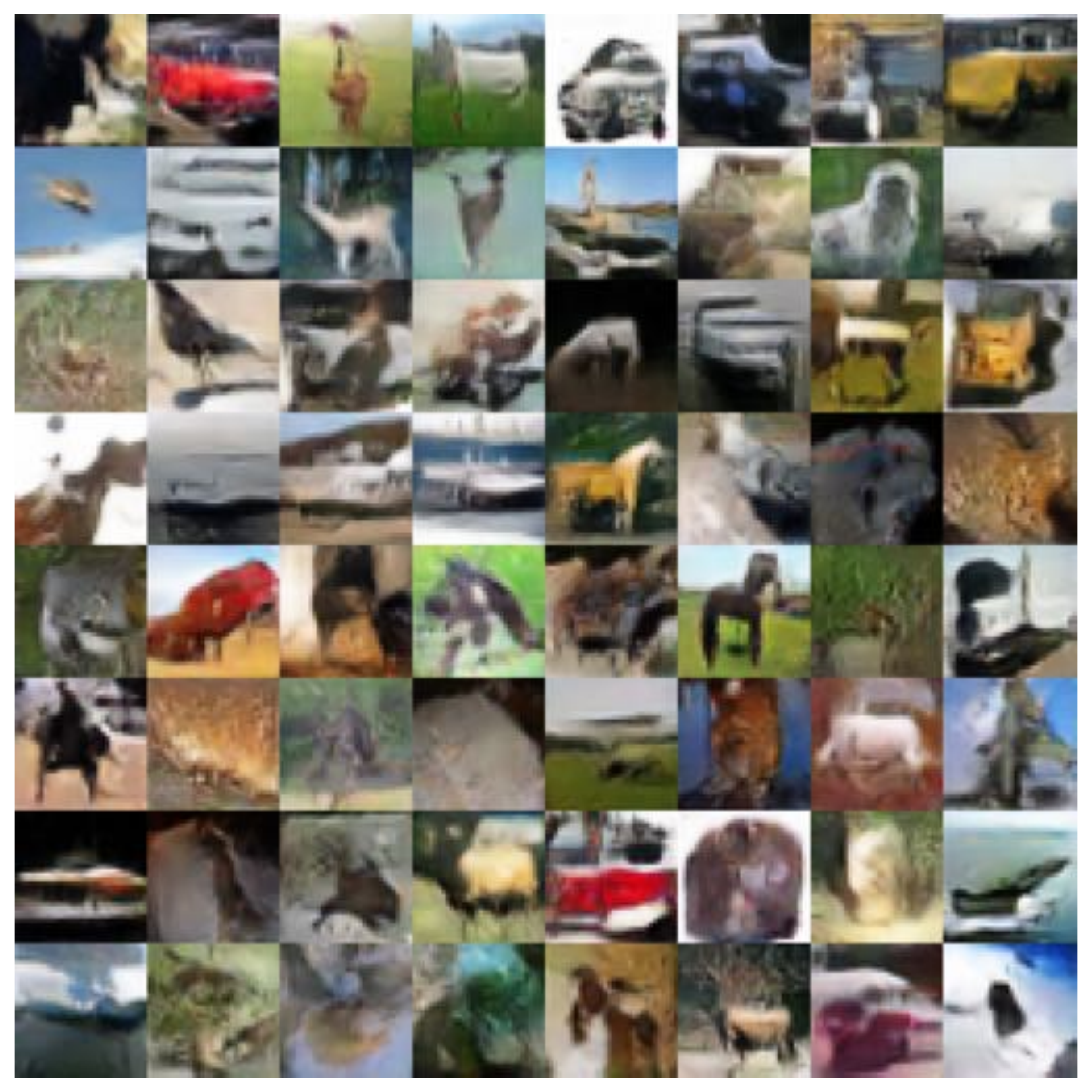}}%
			\subfigure[\rbigan]{\includegraphics[width=0.5\linewidth]{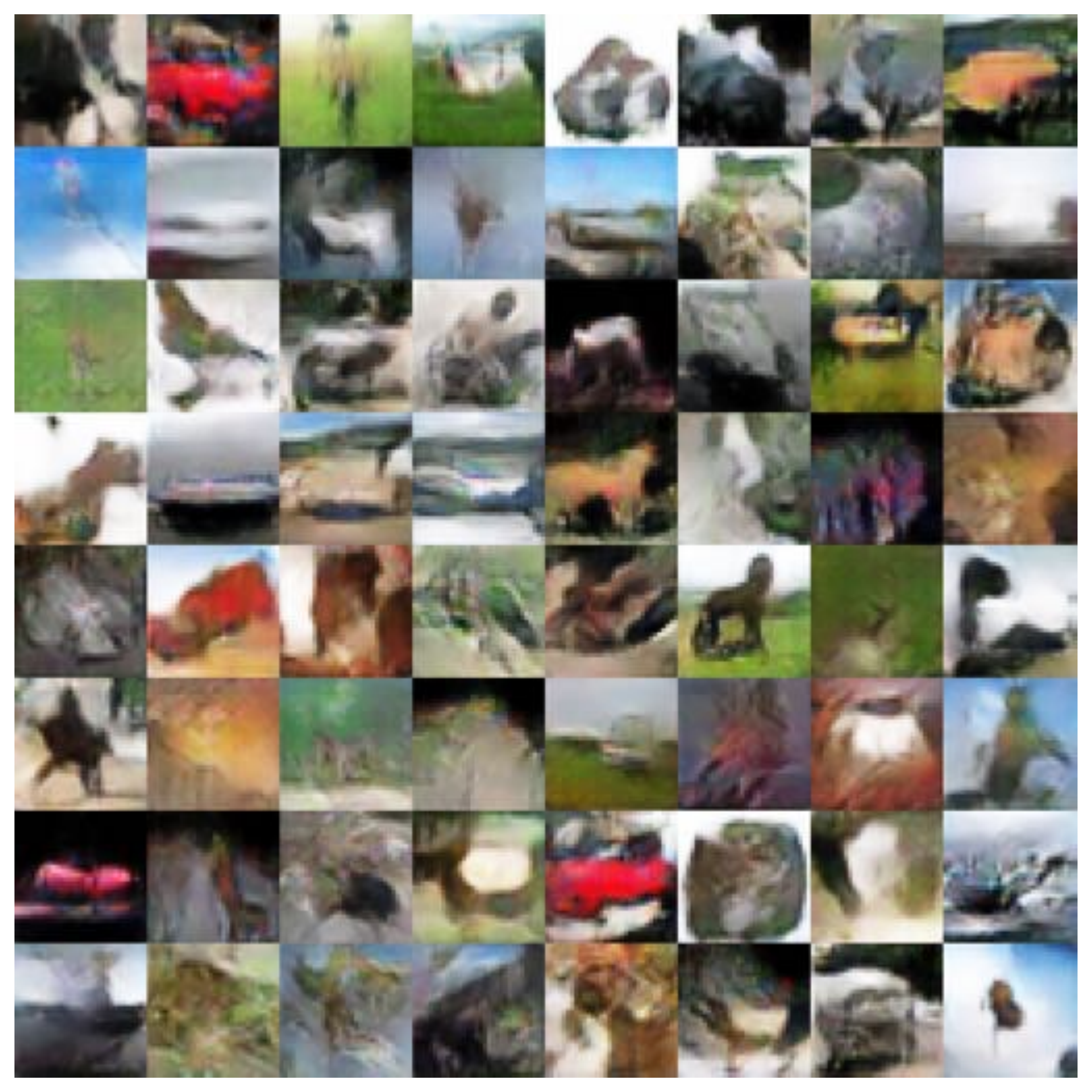}}%
		\end{tabular}
	\end{center}
	\caption{Figure (a) shows samples from the training set of the CIFAR10 dataset. Figures (b), (c) and (d) show the reconstructions of the original samples obtained using \prbiganr, \prbigan\ and \rbigan\ respectively.}%
\label{fig:x_recons_c10}
	\vskip -0.05in
\end{figure*}

\begin{figure*}[h!]
	\begin{center}
		\begin{tabular}{cc}
			
			\subfigure[Orignal]{\includegraphics[width=0.5\linewidth]{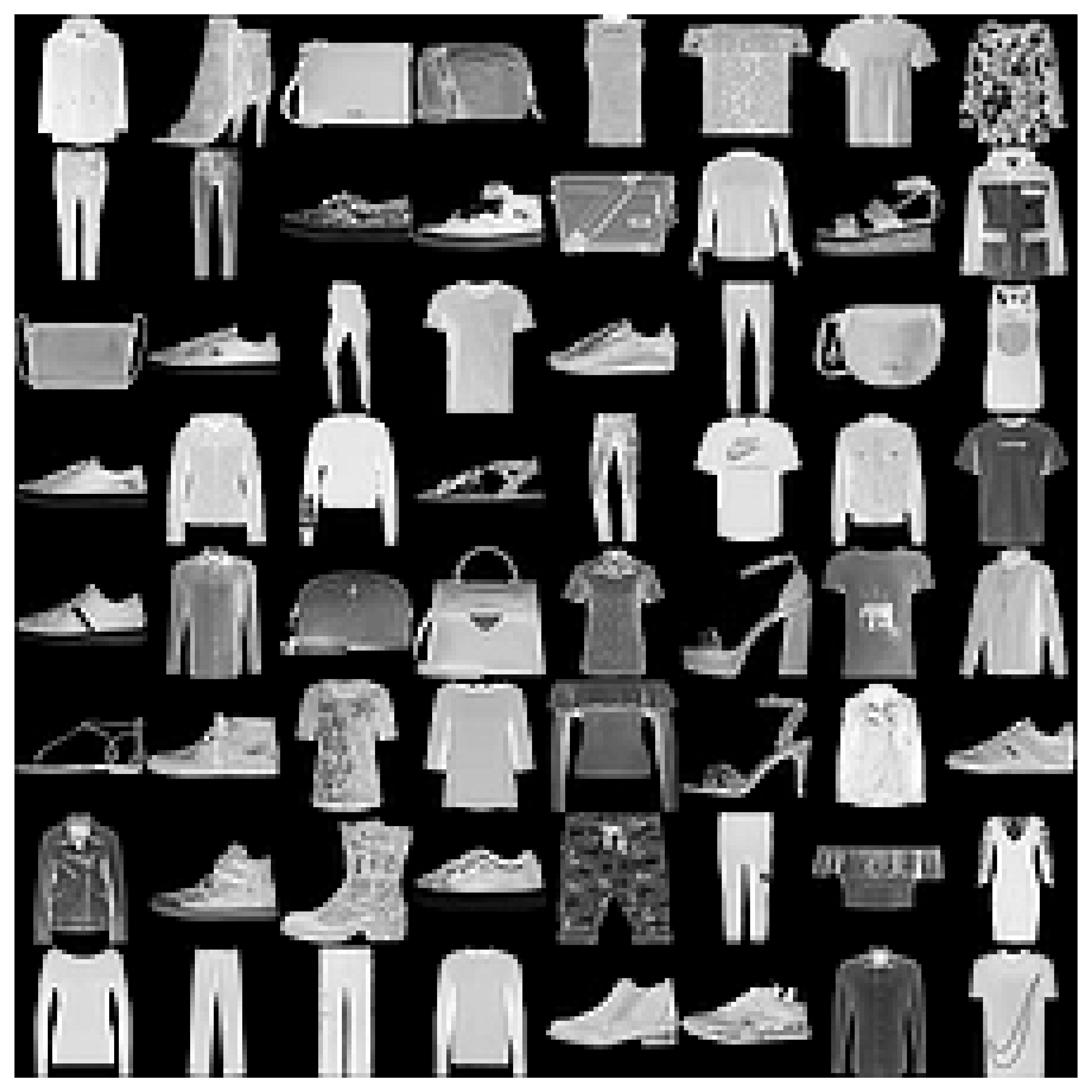}}%
			\subfigure[\prbiganr]{\includegraphics[width=0.5\linewidth]{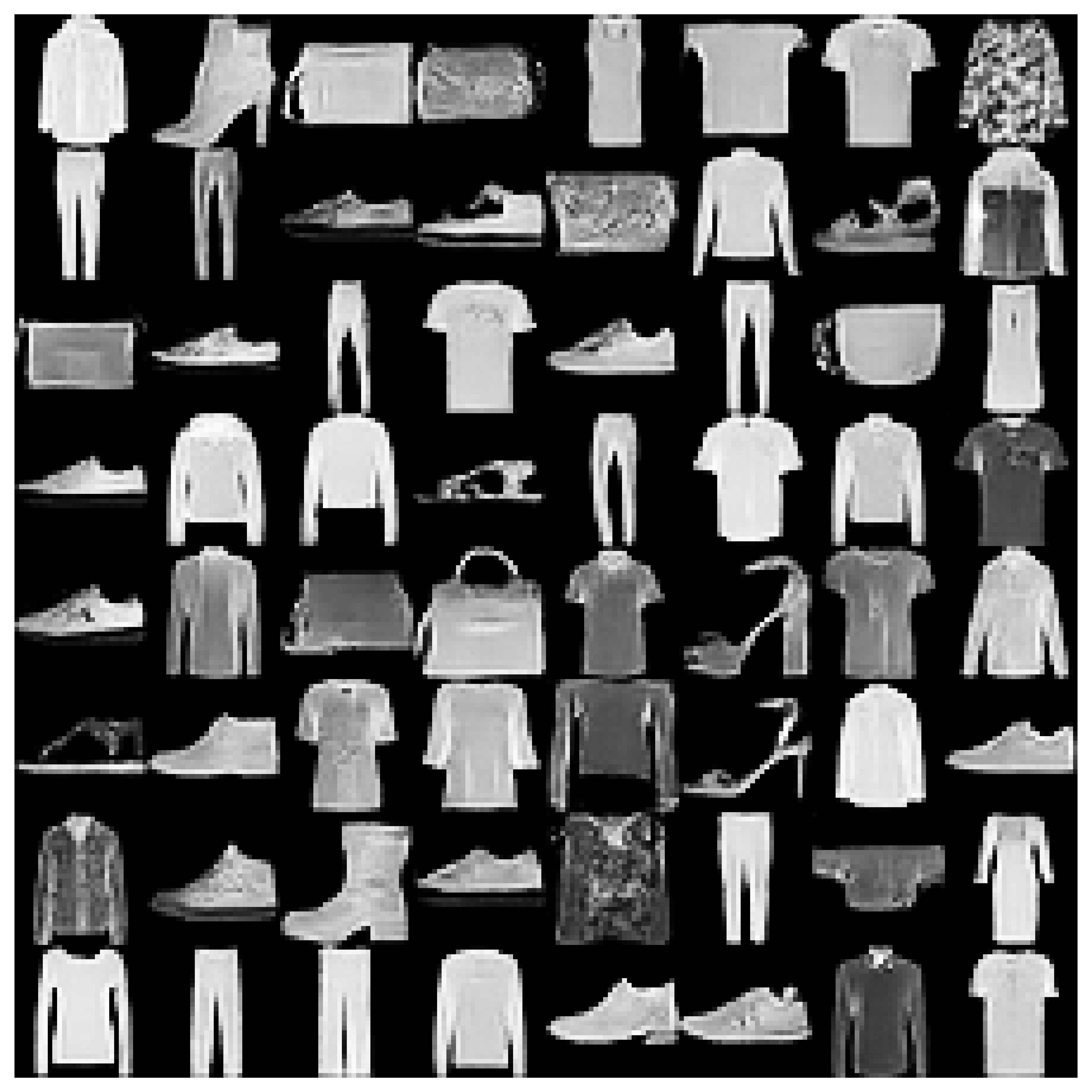}}%
			\\
			\subfigure[\prbigan]{\includegraphics[width=0.5\linewidth]{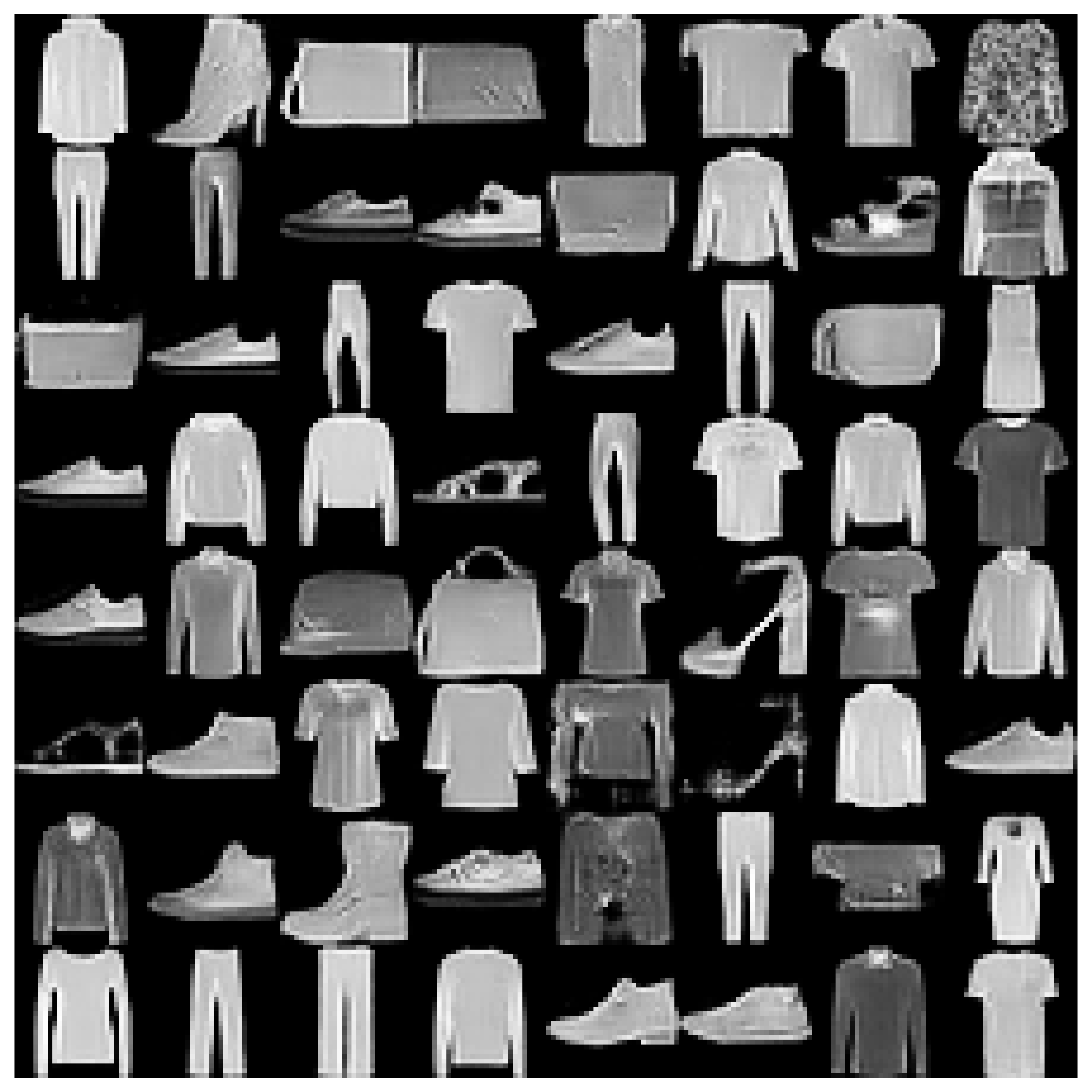}}%
			\subfigure[\rbigan]{\includegraphics[width=0.5\linewidth]{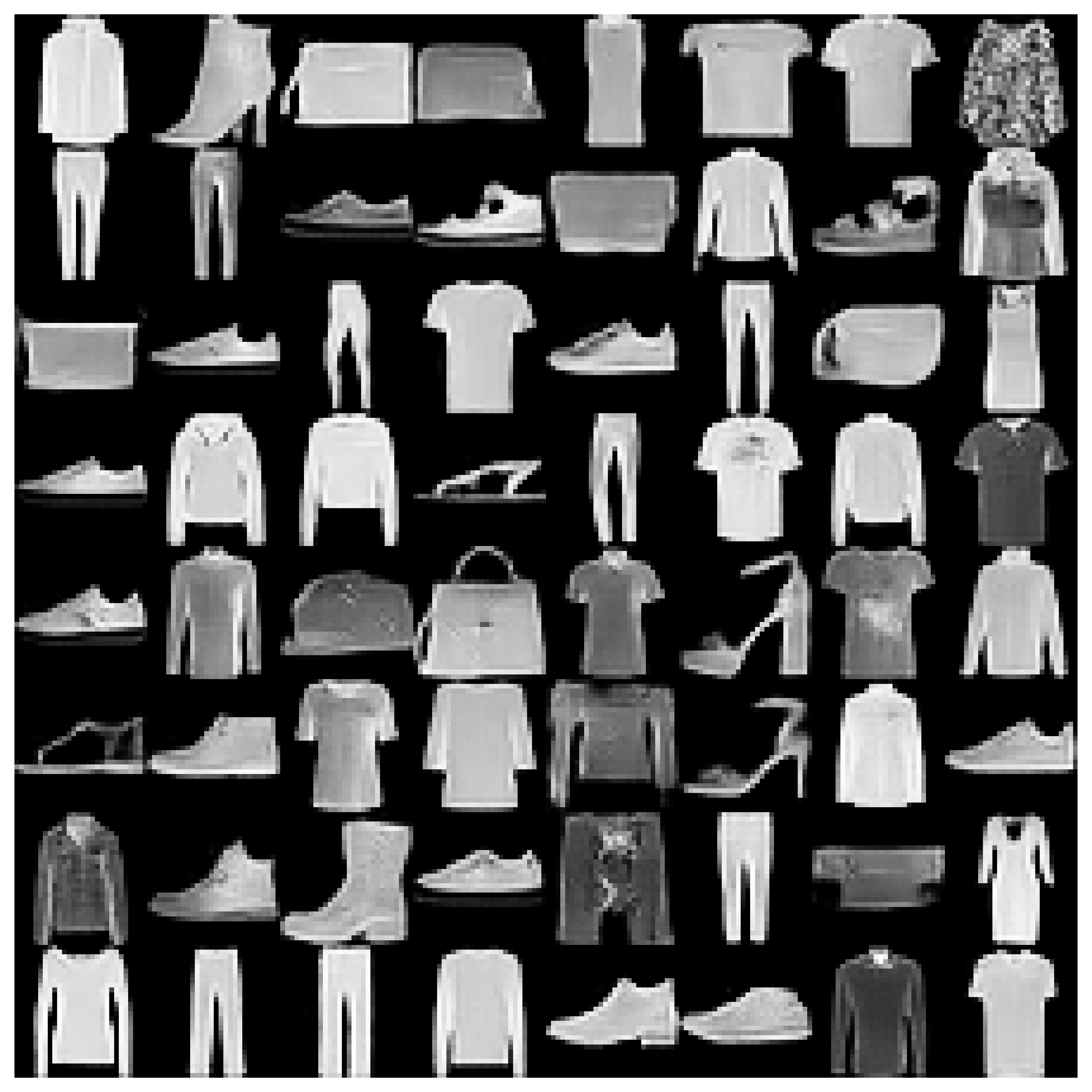}}%
		\end{tabular}
	\end{center}
	\caption{Figure (a) shows samples from the training set of the F-MNIST dataset. Figures (b), (c) and (d) show the reconstructions of the original samples obtained using \prbiganr, \prbigan\ and \rbigan\ respectively.}%
\label{fig:x_recons_fm}
	\vskip -0.05in
\end{figure*}

In Figure \ref{fig:x_recons_c10} we show samples from the CIFAR10 traning set and the corresponding reconstructions obtained using the hyperparameter configurations that achieve the best FID scores of \rbigan, \prbigan\ and \prbiganr, as performed in the main paper. By visual inspection is it difficult to assess which one ccomplish better reconstruction quality. Nonetheless, the value of $\lcyc$ for \prbiganr\ is considerably smaller than for \prbigan\ or \rbigan\ which indicates the reweighting schemes requires less regularization to achieve same performance. This behavior also happens for the F-MNIST as shown in Figure \ref{fig:x_recons_fm}.

\begin{figure*}[h!]
	\begin{center}
		\begin{tabular}{cc}
			
			\subfigure[Orignal]{\includegraphics[width=0.5\linewidth]{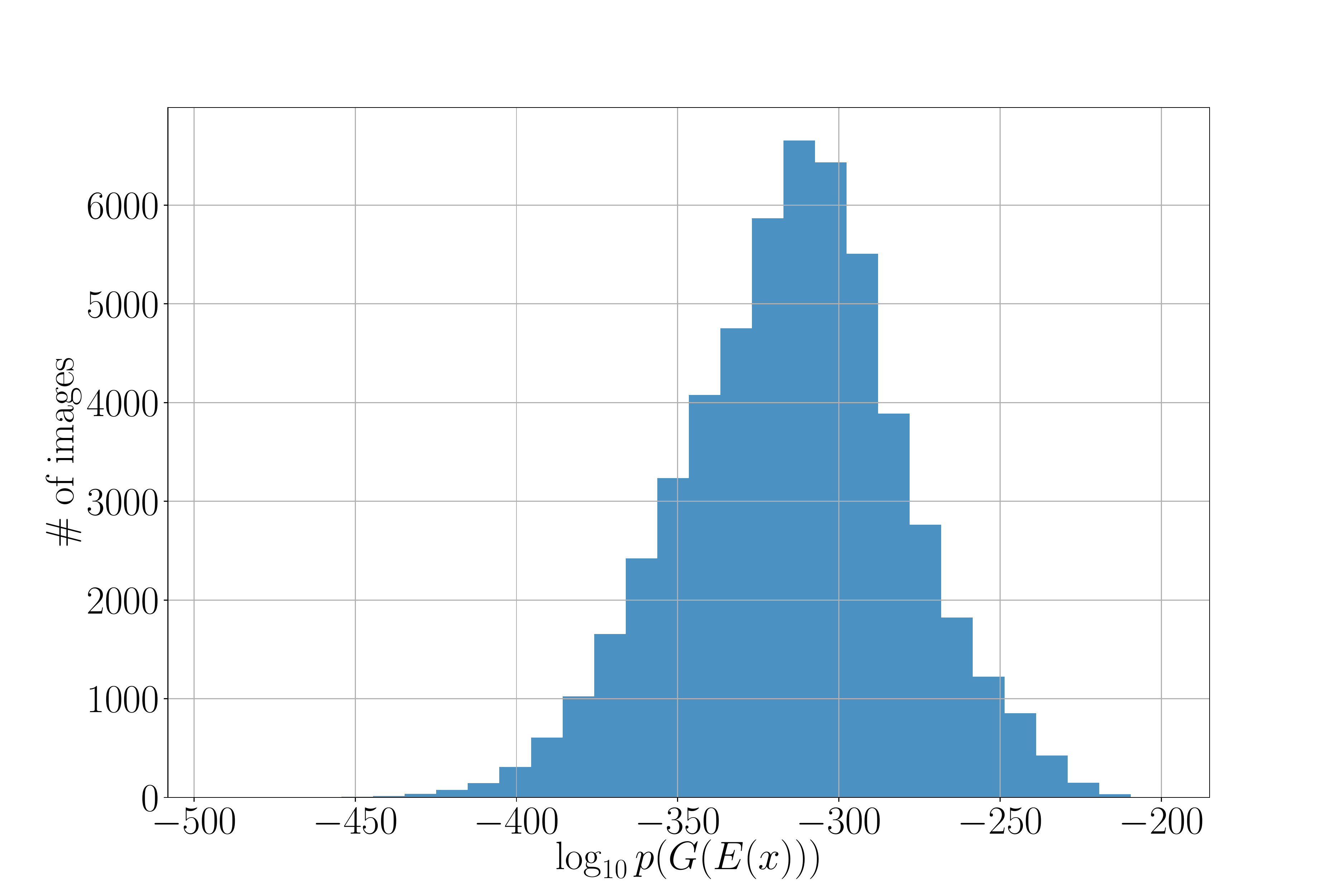}}%
			\subfigure[\prbiganr]{\includegraphics[width=0.5\linewidth]{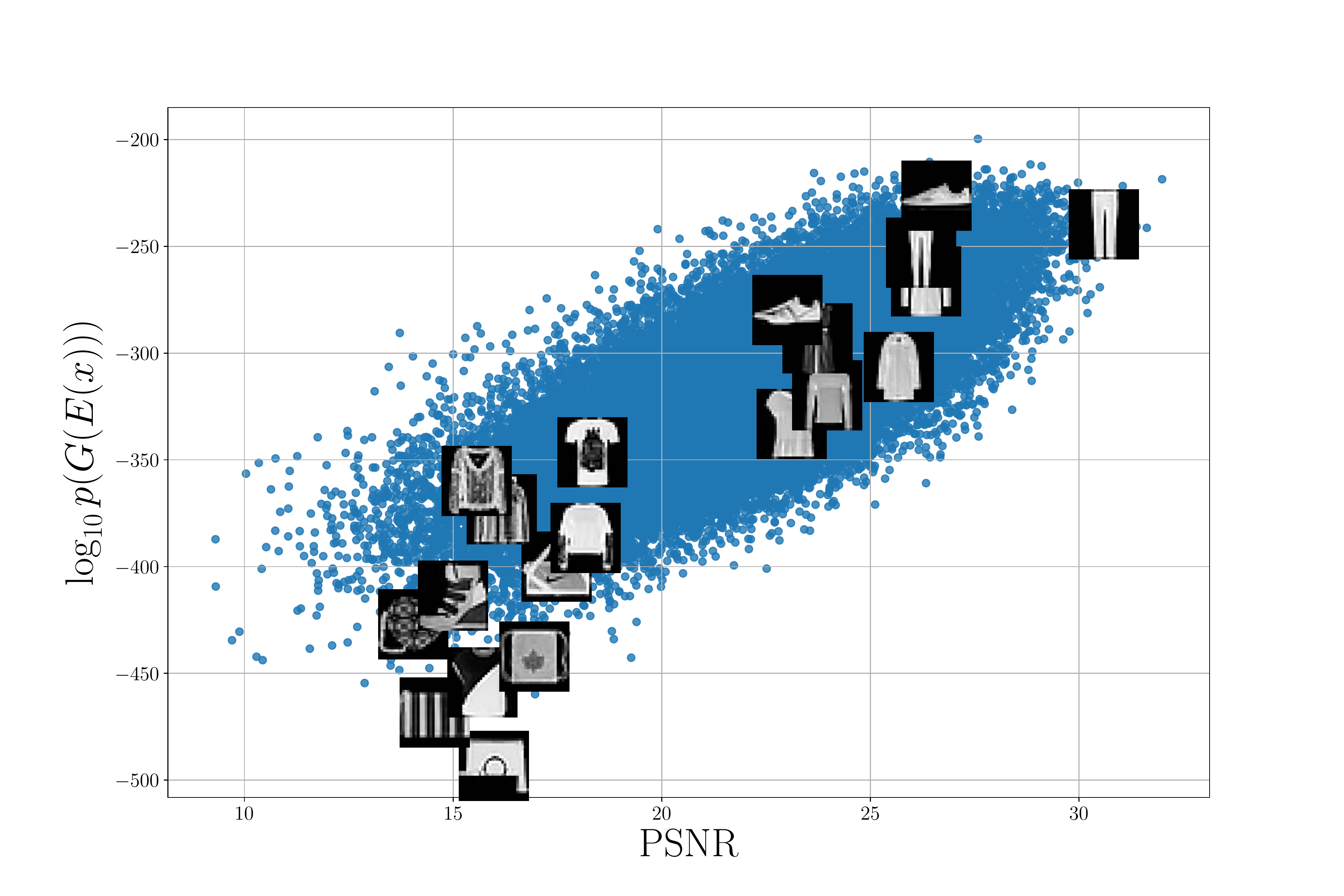}}%
		\end{tabular}
	\end{center}
	\caption{In Figure (a) the empirical distribution of $\logl$ for all training examples. In Figure (b) the scatter plot of reconstruction quality measured in PSNR versus $\logl$ for all examples in the training set with several original images overlapped.  }%
	\label{fig:psnr_improve}
	\vskip -0.05in
\end{figure*}

Figure \ref{fig:psnr_improve} (a) shows the histogram of the marginal log-likelihood of $G(E(\x))$ for all samples in the F-MNIST training set and we can observe the differences are huge, in the order of $10^{100}$. In Figure \ref{fig:psnr_improve} (b) we present a scatter plot of the  reconstruction quality measured in PSNR versus the marginal likelihood. Similarly as with the CIFAR10 dataset, we observe there is a linear relationship between both variables and that simple plain images obtain are more likely and better reconstructed.

\end{document}